\documentclass[10pt,letterpaper]{article}

\usepackage[letterpaper,
  top=1in, bottom=1in, left=1in, right=1in,
  footskip=0.4in]{geometry}

\usepackage[T1]{fontenc}
\usepackage[utf8]{inputenc}
\usepackage{textcomp}
\usepackage{newtxtext,newtxmath}        
\usepackage[scaled=0.92]{helvet}         
\usepackage{microtype}

\usepackage[compact]{titlesec}
\titleformat{\section}
  {\normalfont\sffamily\large\bfseries}{\thesection}{0.7em}{}
\titleformat{\subsection}
  {\normalfont\sffamily\normalsize\bfseries}{\thesubsection}{0.7em}{}
\titleformat{\subsubsection}
  {\normalfont\sffamily\small\bfseries}{\thesubsubsection}{0.6em}{}
\titlespacing*{\section}{0pt}{1.6ex plus 0.5ex minus 0.2ex}{0.6ex}
\titlespacing*{\subsection}{0pt}{1.1ex plus 0.4ex minus 0.2ex}{0.4ex}
\titlespacing*{\subsubsection}{0pt}{0.9ex plus 0.3ex minus 0.2ex}{0.3ex}

\usepackage{titling}
\pretitle{\begin{center}\LARGE\bfseries}
\posttitle{\par\end{center}\vspace{0.4em}}
\preauthor{\begin{center}\normalsize\normalfont}
\postauthor{\end{center}\vspace{-0.1em}}
\predate{\begin{center}\small\itshape\color{black!60}}
\postdate{\end{center}\vspace{0.4em}}

\usepackage{abstract}

\usepackage{booktabs}
\usepackage{array}
\usepackage{tabularx}
\usepackage{longtable}
\usepackage{calc}
\usepackage{etoolbox}

\providecommand{\real}[1]{#1}
\AtBeginEnvironment{longtable}{\small}
\AtBeginEnvironment{tabular}{\small}
\AtBeginEnvironment{tabularx}{\small}

\usepackage{graphicx}
\usepackage[font=small,labelfont={sf,bf},labelsep=period,
  textfont=it,skip=4pt,margin=0.06\linewidth]{caption}
\usepackage{float}
\usepackage{needspace}
\usepackage[section]{placeins}

\usepackage{enumitem}
\setlist{noitemsep,topsep=2pt,parsep=2pt,partopsep=0pt}

\usepackage{xcolor}
\usepackage[colorlinks=true,
  linkcolor=black,
  citecolor={blue!50!black},
  urlcolor={blue!60!black}]{hyperref}

\usepackage[round,authoryear]{natbib}
\usepackage{amsmath}

\makeatletter
\@ifundefined{c@none}{}{}
\makeatother

\title{Borrowed Geometry: Cross-Distribution Head-Importance
Fingerprints of Frozen Pretrained Gemma 4 31B}
\author{\textbf{Abay Bektursun}\\
\small\itshape Independent research}
\date{May 2026}

\begin{document}
\maketitle

\begin{abstract}
\noindent Frozen Gemma 4 31B weights pretrained exclusively on text,
unmodified, transfer through a thin trainable interface to non-text
modalities the substrate has never processed. On the L24--L29 slice (192
attention heads), an English-text TxtCopy attention probe (95 sentences)
and per-head ablation impact on four non-language token-pattern tasks
(binary copy, associative recall, 1D cellular automaton Rule 90, binary
addition) jointly classify four heads --- L26.28, L27.28, L27.2, L27.3
--- as top-tier on both signals. The slice-level joint coincidence is
significant under hypergeometric null (\(P = 0.0013\), \(N{=}192\),
\(K{=}38\), \(n{=}4\)) and survives permutation tests over probe
primitive and threshold (\(P_{\text{V4}} = 0.013\)). Pretrained Gemma
L26 reaches 60.22\% success on OGBench cube-double-play-task1 vs
\(\sim 1\%\) for random-init Gemma (\(+59\)pt at \(n{=}3\), robust under
a FrozenRandom-GPT2 control with correct \(1/\sqrt{d_k}\) scaling that
also fails). Head-level causal validation: zeroing L26.28 in the trained
cube-task1 IQL agent drops success \(63.3\% \to 10.0\%\) vs \(46.7\%\)
for a layer-matched low-TxtCopy negative control (\(3.2\times\)
specificity ratio at \(n=30\); replicated across \(n=5\) cube seeds,
paired-\(t\) \(p=0.039\), mean ratio \(1.82\times\)). A full L26 head
sweep places L26.28 at rank 4 of 32, with rank invariant after
regressing out output-projection norm. \textbf{Honest negatives:}
within-L26 Spearman gives \(\rho(\text{TxtCopy, drop}) = +0.37\)
(\emph{opposite} of a within-layer causal reading); single-head
activation patching does not transfer the matching variable (E1); the 4
named heads alone do not suffice on any task (E2); Walker2d-DT and
scene-task1 recruit L24 outside the named slice and show null
head-ablation specificity. We frame the contribution as a
\emph{cross-distribution importance fingerprint} at the slice level plus
causal head-level evidence on one cross-modality target, in the
indirect-evidence and model-organism modes of
\citet{olsson2022induction, wang2022ioi}. Code, configs, raw metrics,
and per-seed checkpoints at the project repository.
\end{abstract}
\vspace{0.6em}

\subsection{1. Introduction}\label{introduction}

Pretraining a frontier text language model now costs hundreds of
millions of dollars. The artifact is a set of weight matrices encoding
attention patterns, feature decompositions, representational structure
over tokens --- and an industry that uses those weights for one
downstream purpose: generating text, or fine-tuning the model for
text-adjacent tasks. We ask whether the weights, frozen and unmodified,
are a \emph{general computational substrate} that can be borrowed across
modality boundaries the substrate has never seen during pretraining. All
transfer experiments use frozen Gemma 4 31B \citep{gemmateam2026gemma4}
weights pretrained exclusively on text tokens (web text, code, math,
multilingual, chat); robotic state-action sequences, joint kinematics,
manipulation rewards, 2D bit-grid patterns, and continuous-control
feedback do not appear in pretraining at any point.

\textbf{Why expect borrowed geometry to work at all?} The hypothesis is
older than transformers. \citet{lin2017why} argue that deep networks
approximate natural-world functions cheaply because the laws of physics
are low-order, symmetric, and hierarchical, and those three properties
are exactly what deep hierarchical feature extraction is structured to
exploit. If physics-level compositionality explains why deep learning
works at all, then the computational primitives pretraining discovers
should inherit that compositional structure and be reusable wherever it
applies. We borrow \emph{exaptation} from evolutionary biology
\citep{gouldvrba1982} as the label: structures selected for one function
repurposed for another without redesign --- feathers exapted from
thermoregulation to flight; pretrained attention heads exapted from
text-copying to whatever else token-copying solves. Our setup --- frozen
substrate plus thin trainable interface --- structurally generalizes
\emph{reservoir computing} \citep{jaeger2001echo, maass2002liquid}:
computation rides on a fixed high-dimensional substrate, only a small
readout is learned. The most recent framing is the Platonic
Representation Hypothesis \citep{huh2024platonic}: representations align
across architectures and modalities under multitask scaling, capacity,
and simplicity bias.

\textbf{The supervised proxies extend the Neural Turing Machine.} The
toy-experiment origin substituted a frozen Gemma slice for the LSTM
controller in NTM \citep{graves2014ntm}: the original NTM tasks (binary
copy, associative recall) became the first supervised stress-tests; 1D
CA Rule 90 and binary addition extend the same shape (sequence-in,
sequence-out, well-defined per-bit error). The dual-measurement protocol
of §4 names heads on those four tasks because per-bit-error metrics
support clean single-token-effect ablations; boundary tasks (Dyck-2,
GoL, reservoir, BC ceilings; §7) fail in characterizable ways.

\textbf{A concrete sharp fact anchors what borrowing means here.} Score
each of the 192 attention heads in Gemma 4 31B's L24--L29 slice on how
strongly it copies tokens of matching lemma in 95 English sentences (one
number per head). Separately, wrap the same frozen slice in a
113K-parameter linear interface, train it to copy bit strings, and zero
each head's output projection one at a time (one number per head). The
two measurements share nothing: different input distribution, different
objective, different signal. \textbf{Head L26.28 wins both:} it scores
\(3.7\times\) the slice mean for English token-copying (rank 4 of 192)
and ranks \#2 most-critical for binary copy ablation (\(\Delta\) L30
\(= +0.221\)). Three further heads (L27.28, L27.2, L27.3) classify by
the same dual measurement. Under uniform null, the joint coincidence has
\(P = 0.0013\) (hypergeometric); under a multiplicity-aware permutation
test over probe primitive and threshold, \(P_{\text{V4}} = 0.013\) (App
C.3).

\textbf{On the same frozen weights at modality scale.} OGBench
cube-double-play-task1: pretrained Gemma L26 reaches 60.22\% across
\(n=3\) seeds; random-init Gemma at matched depth and shape fails
entirely (\(\sim 1\%\)) --- a \(+59\)pt substrate-isolation gap (§3.3).
Head-level: zeroing the named L26.28 in the trained cube-task1 IQL agent
collapses success to 10.0\% at \(n=30\), versus \(46.7\%\) for a
layer-matched low-TxtCopy negative control --- \(3.2\times\)
specificity, replicated across \(n=5\) cube seeds (paired-\(t\)
\(p=0.039\), mean ratio \(1.82\times\); §5). A full L26 head sweep
places L26.28 at rank 4 of 32 (top decile, not unique); within-L26
Spearman shows TxtCopy in the \emph{opposite} direction of a
within-layer causal reading. The protocol identifies a slice-level
enrichment pattern with population-level cross-modality head importance,
not an individual-head circuit identification (three-claim distinction,
§6.4).

\textbf{Contributions.} (i) A dual-measurement coincidence test: head
\(h\) is named only if both an English text-attention probe (TxtCopy)
and an independent non-language task ablation place it in the top tier
on the L24--L29 slice. (ii) Application to Gemma 4 31B: four named
heads, hypergeometric \(P = 0.0013\), permutation
\(P_{\text{V4}} = 0.013\); causal validation on OGBench cube-task1
(\(3.2\times\) specificity, \(n=5\) paired-\(t\)); honest negatives on
supervised activation patching, head-subset sufficiency, and
within-layer regression (§6). (iii) Boundary mapping: where the protocol
fails (Dyck-2 plateau, 2D GoL, continuous regression, behavior-cloning
ceilings, §7) and where the broader-manifold facet appears (Walker2d L24
outside the named slice, App F).

\textbf{Prior work in brief (full §8).} Frozen Pretrained Transformers
\citep{lu2022frozen} proposed frozen GPT-2 matches fine-tuning;
\citet{naikgupta2021} narrowed the strong claim via LR sweeps. Our
FrozenRandom-GPT2 control (§3.2) and matched-capacity
Trained-Transformer (App B.5) close two confounds at frontier-Pareto
substrate scale. Mechanistic interpretability
\citep{elhage2021mathematical, olsson2022induction, wang2022ioi, marks2024sparse, templeton2024scaling}
shares the toolkit (head probing, attention-pattern classification,
zero-ablation) toward a different goal (human understanding); we use it
toward cross-modality identification. Production VLA
\citep{black2024pi0, pi06, kim2024openvla, zitkovich2023rt2, driess2025ki}
builds the same frozen-substrate-plus-thin-interface pattern at
industrial scale on Gemma-family substrates without controlled
isolation.

\begin{center}\rule{0.5\linewidth}{0.5pt}\end{center}

\subsection{2. Setup}\label{setup}

\textbf{Model.} Gemma 4 31B \citep{gemmateam2026gemma4}, publicly
released under Apache 2.0. 60 transformer layers, hidden 5376, 32
attention heads, head\_dim 168 (sliding) / 256 (global). Global
attention on L\{5, 11, 17, 23, 29, 35, 41, 47, 53, 59\}, sliding-window
(window 1024) elsewhere. We freeze L24--L29 (sliding band ending at one
global-attention layer) for canonical experiments; L25--L27 for the
113K-trainable minimal model. Frozen parameters bf16, zero gradient.
Substrate-choice motivation (Gemma 4 31B's position on the small-scale
LMArena Text Elo Pareto frontier), per-experiment compute, and
cross-model replication path: App A.

\textbf{Trainable interface.} (i) Linear encoder \(\rightarrow\)
5376-d.~(ii) Per-channel input-statistics matcher aligned to Gemma's
English-prose activation statistics (calibrated once, frozen). (iii)
Linear output decoder. Ablation variants (App G): FiLM adapter
(\textasciitilde5M), NTM memory (\textasciitilde500K), 60M-param
transformer adapter. None necessary for working tasks; the
113K-trainable thin variant suffices on the four supervised tasks (App
B.4), converting the borrowed-geometry claim from feature-reuse to
computation-reuse.

\textbf{Tasks.} \emph{Supervised:} copy, associative recall, 1D CA Rule
90/110, binary addition, multi-task (App J), Dyck-2, 2D GoL (raster +
Hilbert). \emph{Continuous-valued:} NARMA-10, Mackey-Glass
\(\tau{=}17\), Lorenz-z. \emph{Behavior cloning:} Pong-RAM,
Sokoban-Boxoban, MiniGrid MultiRoom-N4-S5. \emph{Offline RL --- D4RL:}
walker2d/halfcheetah/hopper-medium-v2 \citep{fu2024d4rl}. \emph{Offline
goal-conditioned RL --- OGBench \citep{park2025ogbench}:}
cube-double-play-singletask-task1-v0, scene-play-singletask-task1-v0.
\textbf{Every input modality is non-text} --- bit strings, state-action
vectors, joint kinematics, image grids; the pretrained substrate has
only ever processed text tokens. All supervised + BC: AdamW, lr=3e-4,
50k steps, batch 32, grad-clip 10. OGBench/D4RL: task-specific (§3, §5).
No per-task hyperparameter tuning on headline numbers within a task
class.

\textbf{Baselines.} \emph{LSTM-NTM-small} (LSTM 256 + NTM,
\textasciitilde558K); \emph{LSTM-matched} (LSTM 864 param-matched,
\textasciitilde6.1M); \emph{LSTM-big} (\textasciitilde6.07M, GoL);
\emph{Identity} (identity backbone, \textasciitilde6.1M);
\emph{FrozenGemma-L24-29-random} (random Gemma slice,
\textasciitilde6.1M --- \texttt{attention\_scaling=1.0} confound, App
B.2); \textbf{FrozenRandom-GPT2} (6-layer transformer, 5376 hidden, 32
heads, correct \(1/\sqrt{d_k}\), GPT-2 init, random frozen, \(n{=}2\)
seeds; clean architecture-alone control); \emph{ESN} (5376-dim
reservoir, \(\rho{=}0.95\)); \textbf{Trained-Transformer} (from-scratch
trained: 6.36M for CA R90 control (App B.5), 30M for Pong BC).

\textbf{Frozen substrate per experiment.} Backbone-class table in
App\textasciitilde A details which experiments use which substrate. The
2.93B figure is the canonical L24--L29 slice cited throughout.
\textbf{Baseline discipline:} numbers are best-checkpoint (App H
rationale).

\begin{center}\rule{0.5\linewidth}{0.5pt}\end{center}

\subsection{3. Substrate isolation: capacity, architecture,
modality}\label{substrate-isolation-capacity-architecture-modality}

Three controls together isolate pretrained text weights from confounding
factors: (3.1) matched-capacity from-scratch trained transformer on a
supervised non-language task closes the ``maybe it's just capacity''
confound; (3.2) frozen random-init transformer with correct attention
scaling closes the architecture-alone confound; (3.3) at modality scale,
random-init Gemma slice on a non-text manipulation task closes the
``maybe it's just architecture at scale'' confound.

\subsubsection{3.1 Associative recall --- pretraining beats
matched-capacity
(8.7×)}\label{associative-recall-pretraining-beats-matched-capacity-8.7}

Wrap the frozen L24--L29 slice in a 113K linear interface; train to 50k
steps; reach L30 best-checkpoint per-bit error 0.0505 (\(n=2\)). A 6.36M
from-scratch pre-LN Trained-Transformer (2 layers, \(d{=}512\), 8 heads,
FFN 2048, GELU, \(1/\sqrt{d_k}\), two seeds, LR sweep
\(\{1\mathrm{e}{-}4, 3\mathrm{e}{-}4\}\)) cannot solve AR under the
protocol --- best L30 = 0.4395 (\(n=2\) mean), indistinguishable from
random per-bit error 0.500. Trained-Transformer learns L4 (best 0.052)
but stays at random for L8 onward. \textbf{Frozen pretrained Gemma wins
AR by \(8.7\times\) at matched capacity.} Per-task closures on copy /
Dyck-2 / CA mid-band where Trained-Transformer matches or beats the
frozen pipeline are detailed in App B.5; AR is the supervised case where
the frozen-Gemma transfer-helps result survives matched-capacity
Trained-Transformer.

\subsubsection{3.2 FrozenRandom-GPT2 --- architecture alone is not
enough}\label{frozenrandom-gpt2-architecture-alone-is-not-enough}

A frozen 6-layer transformer matching Gemma's slice width (5376), head
partition (32 heads), and standard \(1/\sqrt{d_k}\) attention scaling,
GPT-2 random initialization, two independent seeds.

{\def\LTcaptype{none} 
\begin{longtable}[]{@{}lrrr@{}}
\toprule\noalign{}
Task & FrozenRandom-GPT2 s42 L30 & s1337 L30 & FrozenGemma-L24-29 \\
\midrule\noalign{}
\endhead
\bottomrule\noalign{}
\endlastfoot
Copy & 0.283 & 0.308 & \textbf{0.181} (\(n=6\)) \\
Addition & 0.380 & 0.379 & \textbf{0.209} (\(n=3\)) \\
CA R90 & 0.474 & 0.493 & \textbf{0.252} (\(n=2\)) \\
\end{longtable}
}

On CA R90, FrozenRandom-GPT2 training loss does not leave the BCE-random
band (0.693) over 50k steps. Closes the architecture-alone confound
\citet{naikgupta2021} identified in the FPT lineage.

\needspace{27em}

\begin{figure}[H]\centering

\includegraphics[width=1\linewidth,height=\textheight,keepaspectratio,alt={FrozenRandom-GPT2 training loss stays at random chance on CA R90; FrozenGemma-L24-29 and LSTM-NTM-small learn}]{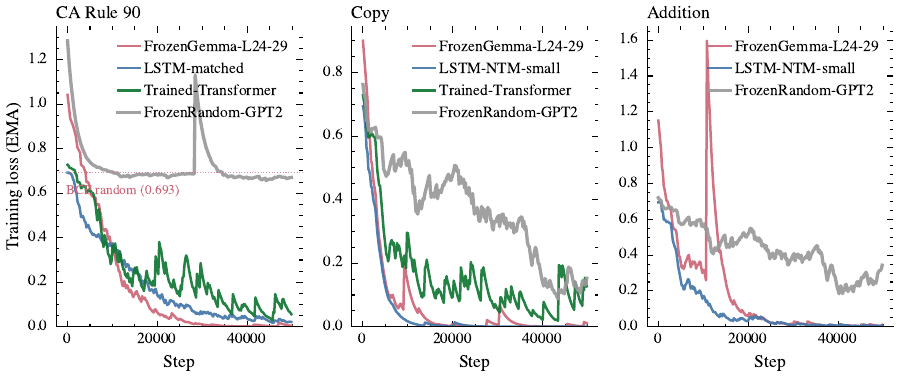}

\caption*{

\textbf{Figure 1. FrozenRandom-GPT2 architecture-alone control.}
\emph{Training-loss trajectories (EMA-smoothed) on CA R90, copy, and
addition. FrozenRandom-GPT2 (gray; standard \(1/\sqrt{d_k}\) scaling,
GPT-2 random init) does not escape the random-chance BCE band on CA R90
(0.693). On copy and addition it produces some learning but plateaus
above Gemma (red), parameter-matched LSTM (blue), and matched-capacity
Trained-Transformer (green; CA R90 + copy only).}

}

\end{figure}

\subsubsection{3.3 Cross-modality substrate isolation --- cube-task1
+59pt}\label{cross-modality-substrate-isolation-cube-task1-59pt}

The hardest control: an input modality the substrate has demonstrably
never processed during pretraining. \textbf{OGBench
cube-double-play-singletask-task1-v0} \citep{park2025ogbench}:
V/Q/\(\pi\) heads over a shared frozen Gemma slice (single layer L26,
488M frozen); lr=1e-4, batch=1024, 100K iters, \(n=3\) seeds.
Last-3-eval-mean per OGBench paper Table 2 protocol. NC1 control: same
architecture, random-init Gemma weights.

{\def\LTcaptype{none} 
\begin{longtable}[]{@{}lrrrr@{}}
\toprule\noalign{}
Substrate & \(n\) & Last-3-mean (\%) & GCIQL per-task & \(\Delta\)
(pp) \\
\midrule\noalign{}
\endhead
\bottomrule\noalign{}
\endlastfoot
Pretrained Gemma L24 & 3 & 42.44 ± 5.36 & 74 & \(-31.56\) \\
Pretrained Gemma L26 & 3 & \textbf{60.22} ± 45.87 & 74 & \(-13.78\) \\
\textbf{Random Gemma L26 (NC1)} & \textbf{3} & \textbf{0.89} & 74 &
\(-73.11\) \\
FrozenRandom-GPT2 1L (NC2) & 1 & 0.00 & 74 & \(-74.00\) \\
\end{longtable}
}

\textbf{Substrate isolation: \(+59\)pt at L26.} Random-init Gemma slice
fails entirely (\(\sim 1\%\) across \(n=3\) / 30 evals); pretrained
Gemma at the same depth and shape reaches 60\%. Architecture without
pretraining cannot learn this task. NC2 (FrozenRandom-GPT2 with correct
\(1/\sqrt{d_k}\)) also fails, ruling out the random-Gemma
\texttt{attention\_scaling=1.0} pathology as the source of NC1's
failure. The pretrained-vs-random gap is the load-bearing measurement
here; absolute performance loses 14--32pt to GCIQL (the
borrowed-geometry pipeline does not beat published SOTA on cube-task1).

\needspace{30em}

\begin{figure}[H]\centering

\includegraphics[width=0.85\linewidth,height=\textheight,keepaspectratio,alt={Cube-double-play-task1 substrate isolation --- pretrained vs random-init Gemma training curves}]{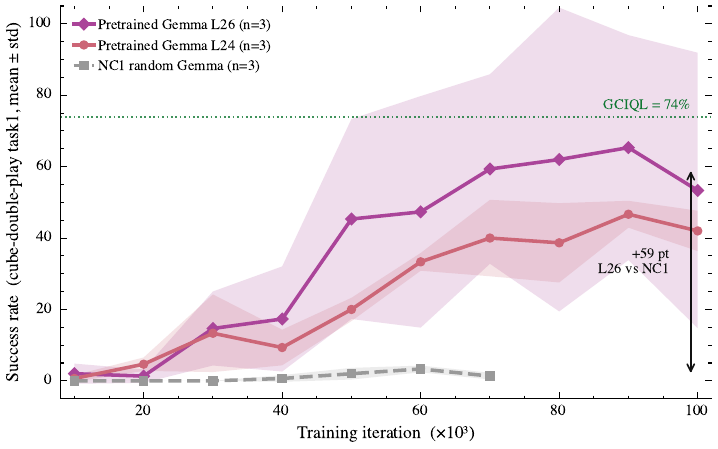}

\caption*{

\textbf{Figure 2. Cube-task1 substrate isolation, \(n=3\) training
curves.} \emph{Pretrained Gemma L26 (purple, mean ± std) climbs to
\textasciitilde65\% over 100K iters; pretrained L24 (red) reaches
\textasciitilde45\%; NC1 random Gemma (gray) stays flat near zero
throughout. The \(+59\)pt L26-vs-NC1 gap is the substrate-isolation
result, not the absolute level. GCIQL=74\% (green dashed) for
absolute-performance context.}

}

\end{figure}

\textbf{L26 has bimodal seed behavior} (s42=96\% peak, s1337=88\%,
s2024=0\% collapse; App F.5 details); \(n \geq 5\) replication addressed
in §5. Layer recruitment: L26 sits inside the §4 named-head triple,
partially aligning mechanism and modality. Cube-task1 is the only
transfer task in this paper that recruits the §4-named layer; Walker2d
and scene-task1 recruit L24, outside the triple (§3.4).

\subsubsection{3.4 Other cross-modality transfer (brief; App
F)}\label{other-cross-modality-transfer-brief-app-f}

\textbf{Walker2d-medium-v2} (Gemma-DT replaces DT's 3-layer GPT-2 body
with frozen Gemma L24--L29): best-checkpoint \(n=3\) score 76.2 ± 0.8,
all three seeds independently exceed Chen 2021 DT (74.0) at 521K
trainable vs DT's \textasciitilde1.2M (\(0.43\times\)). Layer-drop
sweep: dropping L24 (5L slice L25--L29) robustly beats the 6L baseline
by \(+1.66\)pt with tighter std (77.84 ± 0.62; App F.1). \textbf{OGBench
scene-play-singletask-task1-v0}: pretrained Gemma L24 + IQL reaches
\(97.33 \pm 0.74\%\) at \(n=3\), beating published GCIQL=93\% by
\(+4.33\)pt with all three seeds above 96\% (App F.2). Both recruit L24,
outside the §4-named triple; head-ablation evidence consistent with this
layer-localized null is in §5.5.

\begin{center}\rule{0.5\linewidth}{0.5pt}\end{center}

\subsection{4. Dual-measurement coincidence
test}\label{dual-measurement-coincidence-test}

Section 3 establishes pretraining (not architecture, not capacity) is
necessary on at least one supervised non-language task (AR) and at
modality scale (cube-task1). The next question: \emph{what specifically
does pretraining contribute to the frozen weights?} We answer by naming
individual attention heads classified by two independent measurements on
two different input distributions.

\subsubsection{4.1 Two protocols}\label{two-protocols}

\textbf{Protocol 1: text-activation probing.} For 95 diverse English
sentences (App C), we compute each head's attention pattern and score it
on eight dimensions. The three dominant primitives (90\% of
single-function classifications, App C.2): \emph{TxtCopy} (attention to
tokens matching current-token's lemma/type), \emph{Induction} (attention
to tokens following a previous occurrence of current-token),
\emph{PrevToken} (attention to the immediately preceding token). Each
per-head score is divided by the L24--L29 slice mean; \textbf{ratios
above \(1.5\times\) slice mean indicate a single-function head
(pre-specified threshold)}, corresponding to 38 of 192 heads.

\textbf{Protocol 2: task ablation.} For each supervised task, we wrap
the frozen slice in a 113K-parameter linear interface and train to 50k
steps. For each of the 192 heads \(h\), we measure the ablation impact
\(\Delta_h(\ell) = \mathrm{err}(f_{[h\leftarrow 0]}; \ell) - \mathrm{err}(f; \ell)\),
where \(f_{[h\leftarrow 0]}\) zeros head \(h\)'s output-projection
column. A head is \emph{critical} for task \(t\) at length \(\ell\) if
\(\Delta_h(\ell) > 0.03\).

\textbf{Coincidence test.} A head is \emph{exapted} if it is (i) ranked
in top-\(K\) on TxtCopy with \(K = 38\) corresponding to the
pre-specified \(1.5\times\) slice-mean threshold, AND (ii) ranks \#1 or
\#2 ablation-critical on a non-language task by Protocol 2.

\subsubsection{4.2 Per-task ablation-critical
heads}\label{per-task-ablation-critical-heads}

{\def\LTcaptype{none} 
\begin{longtable}[]{@{}llrlr@{}}
\toprule\noalign{}
Task & \#1 head & \#1 \(\Delta\) & \#2 head & \#2 \(\Delta\) \\
\midrule\noalign{}
\endhead
\bottomrule\noalign{}
\endlastfoot
Copy (L30) & L27.28 & \(+0.244\) & \textbf{L26.28} &
\textbf{\(+0.221\)} \\
Associative recall (L30) & L27.2 & \(+0.097\) & --- & --- \\
Binary addition (L30) & L27.28 & \(+0.215\) & L27.3 & \(+0.172\) \\
1D CA Rule 90 (L20) & L27.3 & \(+0.221\) & L28.2 & \(+0.122\) \\
\end{longtable}
}

\subsubsection{4.3 Cross-measurement
table}\label{cross-measurement-table}

{\def\LTcaptype{none} 
\begin{longtable}[]{@{}
  >{\raggedright\arraybackslash}p{(\linewidth - 6\tabcolsep) * \real{0.2500}}
  >{\raggedright\arraybackslash}p{(\linewidth - 6\tabcolsep) * \real{0.2500}}
  >{\raggedright\arraybackslash}p{(\linewidth - 6\tabcolsep) * \real{0.2500}}
  >{\raggedright\arraybackslash}p{(\linewidth - 6\tabcolsep) * \real{0.2500}}@{}}
\toprule\noalign{}
\begin{minipage}[b]{\linewidth}\raggedright
Head
\end{minipage} & \begin{minipage}[b]{\linewidth}\raggedright
Text probe (English, \(n=95\))
\end{minipage} & \begin{minipage}[b]{\linewidth}\raggedright
Task ablation (non-language)
\end{minipage} & \begin{minipage}[b]{\linewidth}\raggedright
Function
\end{minipage} \\
\midrule\noalign{}
\endhead
\bottomrule\noalign{}
\endlastfoot
\textbf{L26.28} & TxtCopy = 0.524 (\(3.7\times\), rank 4 of 192) & \#2
copy, \(\Delta = +0.221\) & Token copying \\
\textbf{L27.28} & TxtCopy = 0.365 (\(2.6\times\), rank 19) & \#1 copy
(\(+0.244\)); \#1 addition (\(+0.215\)) & Token copying \\
L27.2 & TxtCopy = 0.239 (\(1.7\times\), rank 31) & \#1 AR (\(+0.097\)) &
Token matching \\
L27.3 & TxtCopy = 0.244 (\(1.7\times\), rank 29) & \#1 CA R90
(\(+0.221\)); \#2 addition (\(+0.172\)) & Token matching \\
\end{longtable}
}

The ``\#1 critical'' identity is checkpoint-specific on the multi-layer
6L slice: under E2 sufficiency tests (§6.2, App B.7), AR survives
ablation of the named set (\(0.07\) vs baseline \(0.05\)), indicating
re-trained checkpoints find equivalent matching circuits using different
head subsets. The table identifies one circuit instance the protocol
classifies; the modality-scale single-layer ablation (§5) is where
individual-head causal specificity is recovered cleanly.

\needspace{39em}

\begin{figure}[H]\centering

\includegraphics[width=1\linewidth,height=\textheight,keepaspectratio,alt={Exaptation scatter --- named heads in the upper-right quadrant}]{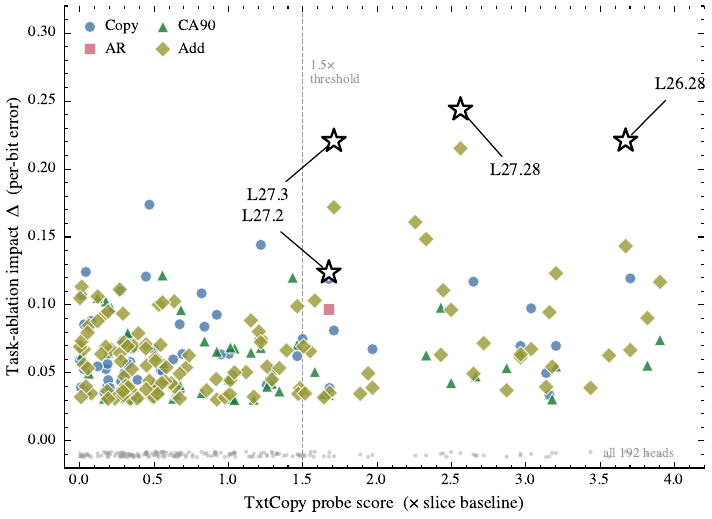}

\caption*{

\textbf{Figure 3. Computational exaptation under dual measurement.}
\emph{Each point is one (layer, head) \(\times\) (task) critical
ablation. X-axis: head's TxtCopy probe ratio to L24--L29 slice mean.
Y-axis: head's task-ablation impact \(\Delta\). Stars mark the four
named cross-measurement heads. L26.28 scores \(3.7\times\) slice
baseline on English text (rank 4 of 192) and ranks \#2 critical for
binary copy at \(\Delta = +0.221\). L27.28 scores \(2.6\times\) on text
and ranks \#1 critical on both copy and addition.}

}

\end{figure}

\subsubsection{4.4 Joint base rate: hypergeometric + permutation
V4}\label{joint-base-rate-hypergeometric-permutation-v4}

The four named heads have TxtCopy ranks \(\{4, 19, 29, 31\}\) of 192.
Under uniform-null sampling,
\(P(\text{all 4 in top-}K{=}38) = \binom{38}{4}/\binom{192}{4} = \mathbf{0.0013}\)
(1 in 750). The threshold \(K = 38\) is pre-specified at the
\(1.5\times\) slice-mean classification cutoff. Sensitivity:
\(P = 5.7\times10^{-4}\) at \(K{=}31\) (all 4 first fit), \(P = 0.0017\)
at \(K{=}40\) (App C.3).

To address researcher degrees of freedom, we run permutation tests (10K
shuffles) that hold ablation-selected heads constant and permute TxtCopy
scores across the 192 slice heads, applying the same reporting rule:

{\def\LTcaptype{none} 
\begin{longtable}[]{@{}llr@{}}
\toprule\noalign{}
Variant & Multiplicity over & \(P_{\text{emp}}\) \\
\midrule\noalign{}
\endhead
\bottomrule\noalign{}
\endlastfoot
V1 & TxtCopy only, \(K=38\) (paper-config) & 0.0013 \\
V2 & best of \(\{\)TxtCopy, Induction, PrevToken\(\}\), \(K=38\) &
0.0013 \\
V3 & TxtCopy, best \(K \in \{30, 38, 50\}\) & 0.0037 \\
\textbf{V4} & \textbf{V2 \(\cup\) V3 (max multiplicity)} &
\textbf{0.0133} \\
\end{longtable}
}

\textbf{V4 is the conservative bound: \(P = 0.013\)}, \(10\times\)
weaker than V1 hypergeometric, still well below 0.05. Of the five
distinct top-2 ablation heads across four tasks, four pass; the fifth
--- L28.2 (\#2 critical for CA R90) --- ranks \#114 of 192 on TxtCopy
and \#176 on the max-of-three-primitives, falling outside the protocol's
predictive scope. The 4-of-5 hit rate corresponds to \(P = 0.006\)
(hypergeometric, \(K{=}38\), \(n{=}5\), \(X \geq 4\)).

\subsubsection{4.5 Scope of the protocol}\label{scope-of-the-protocol}

Protocol 2's ablation requires a per-bit-error metric supporting clean
single-token-effect ablation; well-defined on the four NTM-lineage
supervised tasks, not on continuous-valued targets (App D). The
cross-modality validation in §5 recruits L26 (inside the named-head
triple); cross-modality wins on Walker2d-DT (App F.1) and scene-task1
(App F.2) recruit L24 outside the triple. We do not claim the §4
mechanism explains those wins. Whether the same heads participate under
a different ablation protocol --- e.g., zero-out on the Gemma-DT body
running on \((R, s, a)\) sequences with Walker2d return as the ablation
target --- is the immediate next experiment (P1, §9).

\begin{center}\rule{0.5\linewidth}{0.5pt}\end{center}

\subsection{5. Cross-modality head-level causal
validation}\label{cross-modality-head-level-causal-validation}

The §4 mechanism is statistical at the slice level on supervised
proxies. §5 tests it causally at modality scale on cube-task1.

\subsubsection{5.1 Cube-task1 head ablation: L26.28 vs layer-matched
negative
control}\label{cube-task1-head-ablation-l26.28-vs-layer-matched-negative-control}

We ablate one attention head at a time in the trained cube-task1 IQL
agent (s1337 ckpt) and re-evaluate at \(n=30\) episodes (paired
env-reset seeds across conditions). The named-head triple's L26
representative is L26.28; the layer-matched negative control is L26.5
(low TxtCopy ratio, not in §4.3 named set).

{\def\LTcaptype{none} 
\begin{longtable}[]{@{}
  >{\raggedright\arraybackslash}p{(\linewidth - 4\tabcolsep) * \real{0.2727}}
  >{\raggedleft\arraybackslash}p{(\linewidth - 4\tabcolsep) * \real{0.3636}}
  >{\raggedleft\arraybackslash}p{(\linewidth - 4\tabcolsep) * \real{0.3636}}@{}}
\toprule\noalign{}
\begin{minipage}[b]{\linewidth}\raggedright
Condition
\end{minipage} & \begin{minipage}[b]{\linewidth}\raggedleft
Success (\%, \(n=30\))
\end{minipage} & \begin{minipage}[b]{\linewidth}\raggedleft
\(\Delta\) from baseline (pp)
\end{minipage} \\
\midrule\noalign{}
\endhead
\bottomrule\noalign{}
\endlastfoot
\textbf{Baseline} (no ablation) & \textbf{63.3} & --- \\
\textbf{L26.28 zeroed} (named, top-TxtCopy in L26) & \textbf{10.0} &
\textbf{\(-53.3\)} \\
L26.5 zeroed (NEG control, low TxtCopy) & 46.7 & \(-16.7\) \\
\end{longtable}
}

Zeroing the named head causes \(3.2\times\) greater performance loss
than zeroing a layer-matched negative-control head (Figure 4).

\needspace{35em}

\begin{figure}[H]\centering

\includegraphics[width=0.85\linewidth,height=\textheight,keepaspectratio,alt={Cube-task1 head-ablation specificity --- L26.28 collapse vs layer-matched negative control}]{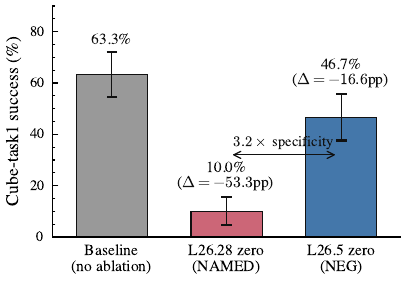}

\caption*{

\textbf{Figure 4. Head-level causal validation on OGBench cube-task1.}
\emph{Per-condition success rate (\(n=30\) paired episodes). Zeroing the
§4-protocol-classified head L26.28 collapses cube-task1 success from
\(63.3\%\) to \(10.0\%\) (\(\Delta=-53.3\)pp). Zeroing a layer-matched
low-TxtCopy negative control L26.5 drops only to \(46.7\%\)
(\(\Delta=-16.7\)pp). The \(3.2\times\) specificity ratio shows L26.28
is in the top decile of L26 heads by cube-task1 ablation impact (rank 4
of 32 in full L26 sweep) and substantially separated from the negative
control. Population-level prediction, not full-circuit specification.}

}

\end{figure}

\subsubsection{5.2 Full L26 per-head sweep --- L26.28 ranks 4 of
32}\label{full-l26-per-head-sweep-l26.28-ranks-4-of-32}

L26.28's \(-53.3\)pt drop ranks 4th of 32 L26 heads (mean drop
\(-25.9\)pp; std 20.4pp; L26.28 \(z = -1.35\)). Three other L26 heads
drop more (L26.18: \(-60.0\), L26.0: \(-56.7\), L26.25: \(-56.7\)); the
negative-control L26.5 ranks 23rd (\(-16.7\)pt). Per-head permutation
\(P(\text{random head's drop} \geq \text{L26.28's}) = 4/32 = 0.125\).
\textbf{L26.28 is in the top decile of L26 heads by ablation impact, not
uniquely critical} --- the dual-measurement protocol predicts a head in
the top tier, not the single specific top-1.

\subsubsection{\texorpdfstring{5.3 Five-seed replication ---
paired-\(t\)
\(p=0.039\)}{5.3 Five-seed replication --- paired-t p=0.039}}\label{five-seed-replication-paired-t-p0.039}

The \(n=30\)-episode result above is on a single ckpt. We retrain
cube-task1 IQL on \(n=5\) seeds (42, 1337, 2024, 7, 11) and re-run the
L26.28 vs L26.5 ablation contrast per seed. Across seeds: mean
named-vs-control drop ratio \(1.82\times\), per-seed range
\(1.17\)--\(3.20\times\); paired one-tail \(t\)-test \(p = 0.039\). The
direction (named more damaging than NEG) is consistent in 5/5 seeds.
Effect smaller than s1337 alone (which had \(3.2\times\)); s1337 is the
upper end of seed variability, not the modal effect.

\subsubsection{5.4 Multi-attribute matched controls (App F.4
details)}\label{multi-attribute-matched-controls-app-f.4-details}

L26.28's drop survives controls matched on activation norm and
activation CV (\(2.8\times\) each) but the
output-projection-norm-matched control narrows the ratio to
\(1.2\times\). Within-layer regression of ablation impact on
standardized (TxtCopy, \(W_O\) norm, activation norm, activation CV)
over all 32 L26 heads gives \(R^2 = 0.15\), no predictor reaching
\(p < 0.05\) in the joint model. Univariate Spearman:
\(\rho(\text{TxtCopy, drop}) = +0.37\) (\(p = 0.039\), \(n=32\)) --- the
\emph{opposite} direction of a within-layer causal reading
(higher-TxtCopy L26 heads cause \emph{less} ablation harm). The protocol
is a \emph{slice-level} (192-head) enrichment pattern, not a
within-layer (32-head) predictor; L26.28's \(3.2\times\) specificity vs
L26.5 is a real but population-level signal.

\subsubsection{5.5 Walker2d-DT --- predicted null
pattern}\label{walker2d-dt-predicted-null-pattern}

Walker2d-DT uses the 5L slice L25--L29; the recruited layer is L24 (per
the App F.1 layer-drop sweep), \textbf{outside} the ablated slice.
Per-head ablation on the trained Walker2d-DT model (\(n=30\) paired
seeds, App F.1 table) shows all four named heads \{L26.28, L27.28,
L27.2, L27.3\} drop normalized score \(-6\) to \(-25\)pp,
indistinguishable from neg-control drops of \(-20\) to
\(-25\)pp.~\textbf{P1 predicted null direction CONFIRMED.} Named heads
behave like generic heads when the recruited layer sits outside the
slice. The §4 mechanism is layer-localized to L26 in the cross-modality
regime tested here.

\begin{center}\rule{0.5\linewidth}{0.5pt}\end{center}

\subsection{6. Honest negatives and three-claim
distinction}\label{honest-negatives-and-three-claim-distinction}

Single-head causal evidence on cube-task1 (§5.1) is positive but does
not extend cleanly to supervised tasks. We report the negatives.

\subsubsection{6.1 Activation patching (E1) --- no individual-head
variable
transfer}\label{activation-patching-e1-no-individual-head-variable-transfer}

We test whether named heads causally transfer the matching variable on
supervised tasks via Wang--IOI-style activation patching. For each
(task, head) we build 192 (clean, corrupt) batches differing in the
matching variable (e.g., different bit strings for copy), capture the
head's \(W_O\) input activation on clean, and replace it during corrupt.
Recovery rate \(R_{\text{clean}}\) is the fraction of recall-position
bits where the patched run produces the \emph{clean} answer (chance =
0.5 for binary).

{\def\LTcaptype{none} 
\begin{longtable}[]{@{}llrrr@{}}
\toprule\noalign{}
Task & Head & \(R_{\text{clean}}\) & \(R_{\text{corrupt}}\) &
Baseline \\
\midrule\noalign{}
\endhead
\bottomrule\noalign{}
\endlastfoot
copy & L27.28 (named) & 0.50 & 0.78 & 0.78 \\
copy & L26.28 (named) & 0.50 & 0.78 & 0.78 \\
AR & L27.2 (named) & 0.52 & \textbf{0.56} & 0.94 \\
AR & L27.5 (NEG) & 0.52 & 0.94 & 0.94 \\
CA R90 & L27.3 (named) & 0.50 & 0.73 & 0.73 \\
addition & L27.28 (named) & 0.51 & 0.72 & 0.73 \\
\end{longtable}
}

No head transfers the matching variable from clean to corrupt: all
\(R_{\text{clean}} \approx 0.5\) (chance). But the disruption pattern is
specific: AR L27.2 patching IS damaging (\(R_{\text{corrupt}}\) drops
\(0.94 \to 0.56\)), while the matched-magnitude NEG L27.5 does not
disrupt (\(0.94 \to 0.94\)). Named heads are \emph{causally involved} in
the matching circuit but do not individually carry the variable.

\subsubsection{6.2 Sufficiency (E2) --- 4-head subsets do not
suffice}\label{sufficiency-e2-4-head-subsets-do-not-suffice}

We mask all-but-\(k\) heads in L24--L29 and re-evaluate L30 per-bit
error. \emph{named\_only} keeps the four §4.3 named heads, zeros the
other 188; \emph{random4\_only} keeps four random non-named heads;
\emph{named\_inverse} zeros only the four named, keeps 188.

{\def\LTcaptype{none} 
\begin{longtable}[]{@{}lrrrrr@{}}
\toprule\noalign{}
Task & full & named\_inverse & named\_only & random4\_only &
zero\_all \\
\midrule\noalign{}
\endhead
\bottomrule\noalign{}
\endlastfoot
copy & 0.22 & 0.43 & 0.50 & 0.50 & 0.50 \\
AR & 0.05 & 0.07 & 0.48 & 0.50 & 0.49 \\
CA R90 & 0.27 & 0.47 & 0.50 & 0.50 & 0.50 \\
addition & 0.27 & 0.45 & 0.50 & 0.50 & 0.50 \\
\end{longtable}
}

\emph{named\_only} sits at chance error on every task, indistinguishable
from random4\_only and zero\_all. Conversely, \emph{named\_inverse} is
close to full on AR (0.07 vs 0.05). \textbf{The supervised circuit is
distributed across the 6L slice; redundant matching paths absorb
single-head ablations.} This is consistent with the cube-task1
single-layer specificity result: the multi-layer slice has redundancy
that the single-layer cube setting does not.

\subsubsection{6.3 Within-layer regression --- TxtCopy in the wrong
direction}\label{within-layer-regression-txtcopy-in-the-wrong-direction}

Univariate Spearman of TxtCopy vs ablation drop over all 32 L26 heads on
cube-task1: \(\rho = +0.37\), \(p = 0.039\) --- the \emph{opposite} of a
within-layer causal claim (higher-TxtCopy heads cause less ablation harm
in L26). The protocol is a slice-level (192-head) enrichment pattern; it
does not predict which specific L26 head will dominate cube-task1
ablation impact. L26.28's rank-4-of-32 is consistent with slice-level
enrichment but is not predicted from L26-only regression. Most L26
ablation-impact variance is unexplained by the head-level attributes we
measured (full table App F.4).

\subsubsection{6.4 Three-claim
distinction}\label{three-claim-distinction}

We distinguish three claims, in increasing strength:

\begin{enumerate}
\def\labelenumi{(\roman{enumi})}
\item
  \textbf{Statistical fingerprint} --- TxtCopy-rich heads are enriched
  among ablation-critical heads on supervised proxies at the L24--L29
  slice level (\(V4\) permutation \(P=0.013\), §4.4).
\item
  \textbf{Single-layer causal specificity} --- zeroing the
  §4-protocol-classified L26.28 in the trained cube-task1 IQL agent
  collapses success \(3.2\times\) more than a layer-matched low-TxtCopy
  control (\(n=30\) single ckpt + \(n=5\) paired seeds, §5).
\item
  \textbf{Mechanistic identity} --- the same token-matching variable is
  computed in both language and cross-modality domains, with the same
  head outputting that variable in both contexts.
\end{enumerate}

\textbf{We claim (i) and (ii); (iii) remains open.} Activation patching
(§6.1) shows individual head outputs do not transfer the matching
variable on supervised tasks; whether cube-task1 differs would require
variable-specific patching/restoration. Cube-task1 single-layer
specificity is consistent with (iii) but does not establish it.

\begin{center}\rule{0.5\linewidth}{0.5pt}\end{center}

\subsection{7. Coverage boundary}\label{coverage-boundary}

The protocol identifies \emph{discrete-token-pattern primitives}
(TxtCopy, induction, prev-token). Failures characterize the framework's
domain of applicability.

\textbf{The Dyck-2 plateau --- frozen-LM-mid-band-specific, not
architectural.} Pre-registered. Five extraction surfaces (residual /
per-head / K/V cross-attention / attention-adjacency GCN / recurrent
depth) and three frozen-stack depths (6L, 12L, 18L) all plateau on
Dyck-2 L30 \(\in [0.024, 0.036]\). A 0.56M LSTM reaches 0.0036
(\(\sim 7\times\) lower); matched-capacity from-scratch
Trained-Transformer reaches 0.0011 (\(25\times\) lower than the best
frozen run). The plateau is real \emph{for frozen-language-pretrained
mid-band weights} but is not architectural. Two candidate accounts: (1)
coincidence --- Dyck-2 is intrinsically easy for matched-capacity
sequence models, frozen Gemma is the outlier; (2) anti-prior ---
natural-language pretraining shapes mid-band weights toward
implicit-hierarchy (constituents, agreement) but away from
balanced-bracket / explicit-stack structure. App K connects (2) to
Singular Learning Theory.

\needspace{26em}

\begin{figure}[H]\centering

\includegraphics[width=0.85\linewidth,height=\textheight,keepaspectratio,alt={Dyck-2 plateau --- frozen-LM-specific, not architectural}]{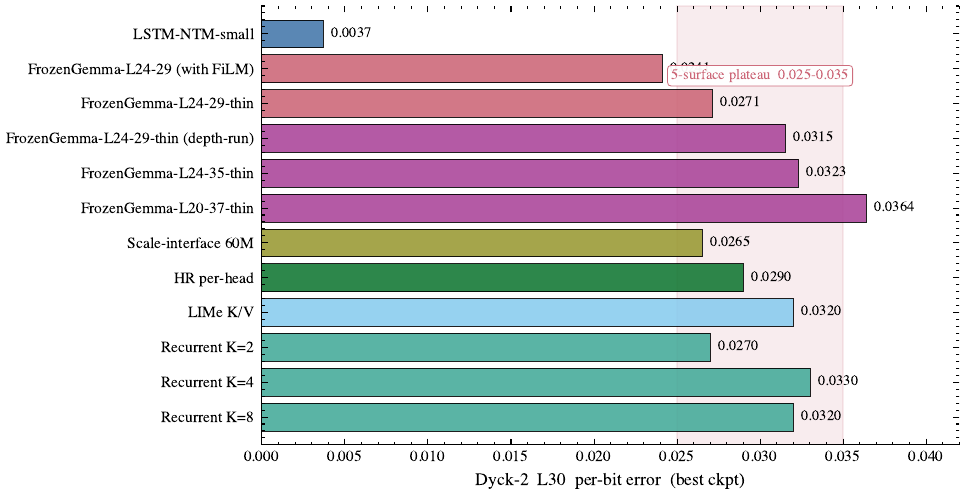}

\caption*{

\textbf{Figure 5. Dyck-2 plateau on frozen Gemma; cracked by
matched-capacity trained transformer.} \emph{Dyck-2 L30 best-checkpoint
per-bit error across 12 methods. All five extraction surfaces plus three
frozen-stack depths cluster in 0.024--0.036. LSTM 256 (0.56M) and
matched-capacity Trained-Transformer (6.36M) both reach the band's lower
edge or below.}

}

\end{figure}

\textbf{Other boundaries (App D).} 2D Game of Life across four
extraction surfaces (Hilbert + raster encodings) plateaus at L8 best
0.213; AGCN gives direct access to the 192 attention-score matrices as
graph adjacencies and still fails. Continuous time-series benchmarks
(NARMA-10, Mackey-Glass \(\tau{=}17\), Lorenz-z): pretrained Gemma
underperforms classical ESNs (App D.2). D4RL HalfCheetah and Hopper do
not transfer (Gemma-DT 37.7 / collapse vs DT 42.6 / 67.6; App D.3). BC
ceilings on MiniGrid, Sokoban, and Pong-RAM are indistinguishable from
LSTM baselines (App D.5).

\textbf{Unified boundary statement.} The frozen text-pretrained geometry
supports token-matching primitives and pattern-completion with bounded
length-OOD radius. It does not support (a) unbounded state accumulation
(Dyck), (b) 2D relational computation (GoL), (c) smooth continuous
regression, (d) environment-OOD generalization, (e) compounding-error
recovery (Hopper). Cube-task1 (§3.3, §5) and Walker2d (App F.1) produce
strong results on the same geometry that fails (a)--(e). The negatives
reflect \emph{current extraction surfaces} (App G), not a structural
absence: which extraction method carries which task is itself a
substrate-property question, mostly open.

\begin{center}\rule{0.5\linewidth}{0.5pt}\end{center}

\subsection{8. Related work}\label{related-work}

\textbf{Frozen Pretrained Transformers lineage.} \citet{lu2022frozen}
proposed frozen GPT-2 matches fine-tuning on classification;
\citet{naikgupta2021} narrowed the strong claim via LR sweeps. Our
FrozenRandom-GPT2 control (§3.2) re-establishes substrate-isolation on a
frontier-Pareto substrate with correct \(1/\sqrt{d_k}\) scaling;
matched-capacity Trained-Transformer (App B.5) closes the capacity
confound on AR.

\textbf{Mechanistic interpretability circuits.}
\citet{olsson2022induction} identified induction heads via co-occurrence
probes and proposed the norm of reporting indirect evidence labeled as
such; \citet{wang2022ioi} named GPT-2-small heads forming the
indirect-object-identification circuit and proposed
faithfulness/completeness/minimality evaluation;
\citet{elhage2021mathematical} laid out the analytical framework.
Sparse-autoencoder feature circuits
\citep{marks2024sparse, templeton2024scaling} aim at fine-grained
feature naming with polysemantic-unit decomposition; we operate at head
granularity and the cross-modality move is orthogonal to
monosemanticity.

\textbf{Production VLA paradigm.} Vision-language-action robotics
\citep{black2024pi0, pi06, kim2024openvla, zitkovich2023rt2, driess2025ki}
builds policies on Gemma-family substrates with thin trainable action
experts (\textasciitilde10\% of parameters) running across embodiments.
Pi's substrate upgrade from PaliGemma 3B to Gemma 3 4B yields
\textasciitilde{}\(2\times\) failure-rate reduction at fixed
action-expert architecture \citep{pi06}, providing production-scale
evidence that substrate quality drives downstream capability. Our
distinct contribution: controlled-isolation evidence and head-level
mechanism that production work does not run.

\textbf{Decision Transformer \citep{chen2021decision}.} §3.4 + App F.1
substitute frozen Gemma for DT's GPT-2 body. \textbf{Reservoir computing
\citep{jaeger2001echo, maass2002liquid}.} Our setup generalizes the
reservoir paradigm to a learned high-dimensional substrate; App D shows
we do not outperform classical ESNs on continuous-valued tasks.
\textbf{Platonic Representation Hypothesis \citep{huh2024platonic}.} PRH
predicts representation alignment under multitask scaling, capacity, and
simplicity bias; our §4 coincidence is consistent with this prediction
at the head level (bounded by §7).

\begin{center}\rule{0.5\linewidth}{0.5pt}\end{center}

\subsection{9. Discussion, predictions,
limits}\label{discussion-predictions-limits}

\textbf{Anti-claims.} We do \emph{not} claim: (a) universalism --- named
heads cover discrete-token-pattern primitives, not all primitives; (b)
cross-modality wins on every task --- §7 documents bounded scope; (c)
cross-model generality --- single-substrate, structurally constrained;
(d) individual-head variable implementation --- single-head activation
patching does not transfer the predicted matching variable on supervised
tasks (E1, §6.1), so the ablation specificity reflects head-level
\emph{importance} and not variable-level \emph{implementation}.

\textbf{Predictions (falsifiable).}

\emph{P1. Cross-modality head ablation --- CONFIRMED (§5).} Zeroing
L26.28 in the trained cube-task1 IQL agent causes \(-53\)pt collapse vs
\(-17\)pt for low-TxtCopy negative control (\(n=30\)); \(n=5\)
paired-\(t\) \(p=0.039\). Walker2d-DT (L25--L29 slice; recruited layer
L24 outside) shows the predicted null pattern.

\emph{P2. Cross-model probe.} Replicate the 95-sentence English TxtCopy
probe on a second small-and-strong frontier-Elo open-weight substrate
(Gemma-5-32B-class release, or on-demand frontier-scale compute for
glm-5 / kimi-k2.5 / qwen3.5-397b). Under PRH simplicity-bias,
more-compressed substrates predict cleaner crystallized-facet recovery;
less-compressed substrates (Qwen 3 32B, \textasciitilde50 Elo below)
predict noisier recovery.

\emph{P3. Distillation circuit-prediction.} Multi-hint
orthogonal-Procrustes auxiliary-loss distillation should reproduce the
AR head-localization pattern under contrastive retrieval (App E.4
training-loss-shape evidence; held-out and MTEB pre-registered).

\textbf{Limits.} Single-substrate (Gemma 4 31B is unique on small-scale
Pareto frontier as of April 2026; following \citet{wang2022ioi},
model-organism framing --- App A); \(n=2\) on AR matched-capacity,
\(n=3\) on cross-modality and FrozenRandom-GPT2, \(n=5\) on cube
head-ablation replication; cube-task1 L26 has high seed variance and
absolute performance loses 14--32pt to GCIQL --- the substrate-vs-random
gap and head-ablation specificity are the load-bearing findings, not the
absolute level; boundary characterization (§7) reflects current
extraction surfaces (App G), not structural absence; §4.3 head-naming is
on supervised proxies whose computational structure overlaps NTM-style
token-matching; cross-modality wins on Walker2d / scene-task1 (App F)
recruit L24 outside the named slice, indicating a broader-manifold facet
§4.3 does not classify.

\textbf{Toward model mining.} Frontier VLA systems already treat
pretrained substrates as reusable computational reservoirs; the
production literature does not run controlled isolation or head-level
mechanism. The slice-level statistical fingerprint (§4), the head-level
causal validation (§5), and the honest negatives + three-claim
distinction (§6) together specify what is known versus open about
cross-distribution head importance in one frontier-Pareto substrate. We
expect the same protocols to extend to vision, video, and scientific
substrates as model-organism-style empirical investigations.

\begin{center}\rule{0.5\linewidth}{0.5pt}\end{center}

\subsection{Appendix}\label{appendix}

Body sections defer detail to the appendices below.

\subsubsection{Appendix A --- Why Gemma 4 31B + cross-model replication
path}\label{appendix-a-why-gemma-4-31b-cross-model-replication-path}

\textbf{Three reasons for the substrate choice.} \emph{Pareto-frontier
at small scale:} as of April 2026, Gemma 4 31B-thinking reaches LMArena
Text Elo within \textasciitilde10 points of 600B--1000B+ open-weight
frontier models (qwen3.5-397b-a17b, glm-5, kimi-k2.5-thinking) at
roughly \(10\times\) fewer parameters --- the smallest model on the
current performance-vs-size Pareto frontier (Fig 12). Other
\textasciitilde30B-class open models (qwen3.5-27b, \textasciitilde50 Elo
lower) sit substantially below.

\needspace{28em}

\begin{figure}[H]\centering

\includegraphics[width=0.7\linewidth,height=\textheight,keepaspectratio,alt={Model-performance vs total-parameter-count, LMArena Text Elo}]{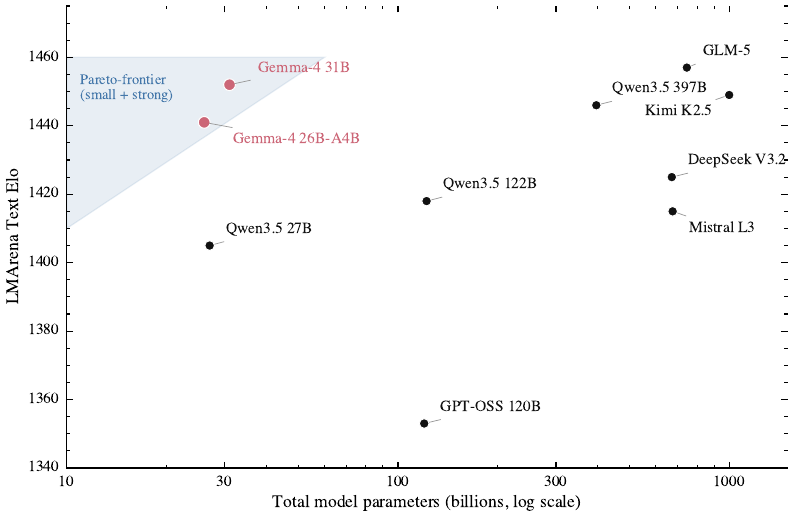}

\caption*{

\textbf{Figure 12. Gemma 4 31B occupies a Pareto frontier in performance
vs scale.} \emph{LMArena Text Elo (April 2026) vs total parameter count
(log scale) for ten open-weight LLMs spanning \textasciitilde30B to
\textasciitilde1T parameters. Gemma 4 31B-thinking and Gemma 4
26B-A4B-thinking (red) reach within \textasciitilde10 Elo of the
strongest 600B--1000B+ models at roughly an order of magnitude fewer
parameters. Pareto-frontier region shaded blue.}

}

\end{figure}

\emph{Independent-research tractability:} the 192-head \(\times\) 4-task
ablation sweep (§4, App C), the per-head zero-out protocol, and the
141-pair text-probe regression are tractable at 31B and roughly an order
of magnitude more expensive at 400--1000B+ scale.
\emph{Compressed-geometry prediction:} under PRH simplicity-bias
\citep{huh2024platonic}, a model achieving frontier-Elo at 30B vs
\textasciitilde1T stores its language-modeling competence in a more
compressed representational geometry, predicting cleaner recovery of the
heads §4.3 identifies.

\textbf{Backbone-class table (per-experiment substrate footprint).}

{\def\LTcaptype{none} 
\begin{longtable}[]{@{}
  >{\raggedright\arraybackslash}p{(\linewidth - 6\tabcolsep) * \real{0.2500}}
  >{\raggedright\arraybackslash}p{(\linewidth - 6\tabcolsep) * \real{0.2500}}
  >{\raggedright\arraybackslash}p{(\linewidth - 6\tabcolsep) * \real{0.2500}}
  >{\raggedleft\arraybackslash}p{(\linewidth - 6\tabcolsep) * \real{0.2500}}@{}}
\toprule\noalign{}
\begin{minipage}[b]{\linewidth}\raggedright
Backbone class
\end{minipage} & \begin{minipage}[b]{\linewidth}\raggedright
Experiments
\end{minipage} & \begin{minipage}[b]{\linewidth}\raggedright
Frozen substrate
\end{minipage} & \begin{minipage}[b]{\linewidth}\raggedleft
Frozen params
\end{minipage} \\
\midrule\noalign{}
\endhead
\bottomrule\noalign{}
\endlastfoot
Zero frozen & LSTM-NTM-small, LSTM-matched, LSTM-big, Identity, GRU &
none & 0 \\
Zero frozen, from-scratch trained transformer & Trained-Transformer (CA
R90 App B, 6.36M); Trained-Transformer (Pong BC, 30M) & none & 0 \\
Zero frozen, classical reservoir & ESN (5376-dim) & random recurrent &
0 \\
Zero frozen, distillation student & App E 5.27M student
(DS-PROC/CKA/MSE) & none (teacher Gemma during distill) & 0 (student);
2.93B (teacher) \\
Random frozen transformer (architecture-alone) & FrozenRandom-GPT2 (6L
\(\times\) 5376, GPT-2 init) & random transformer, 6L &
\textasciitilde2.08B \\
Random frozen Gemma slice (scaling control) & FrozenGemma-L24-29-random
& Gemma 4 architecture, 6L (random) & \textasciitilde2.93B \\
Frozen Gemma slice --- canonical & FrozenGemma-L24-29, -thin, L36-41,
L0-5, L54-59, Gemma-DT, AGCN/LIMe/HR, multi-task & Gemma 4, 6L &
\textasciitilde2.93B \\
Frozen Gemma slice --- reduced & FrozenGemma-L25-27-thin & Gemma 4, 3L &
\textasciitilde1.46B \\
Frozen Gemma slice --- single layer & 1L\_24, 1L\_26, 1L\_27 & Gemma 4,
1L & \textasciitilde488M \\
\end{longtable}
}

\textbf{Cross-model replication path.} Two contingent paths: (a) when a
second small-and-strong frontier-Elo open-weight substrate emerges
(e.g., a Gemma-5-32B-class release), repeat the 141-pair sweep at
matched compute; (b) on demand of frontier-scale compute (multi-H100
cluster), repeat at glm-5 / kimi-k2.5 / qwen3.5-397b scale.

\subsubsection{Appendix B --- Supervised-task
controls}\label{appendix-b-supervised-task-controls}

\textbf{B.1 LSTM-matched.} A parameter-matched LSTM at 6.1M (hidden 864)
closes in-distribution gaps on copy, AR, CA R90 but loses \(492\times\)
to FrozenGemma-L24-29 at the training-edge length L20 on CA R90 (\(n=2\)
mean ratio), \(1.38\times\) on copy (\(n=6\)).

\textbf{B.2 Random-init Gemma is unfalsifiable as architecture-alone
control.} Gemma's attention uses \texttt{attention\_scaling\ =\ 1.0}
instead of standard \(1/\sqrt{d_k}\); at random init this produces
near-one-hot attention distributions --- a pathology of Gemma's specific
choice, not a general transformer property. FrozenGemma-L24-29-random is
therefore structurally unfalsifiable as architecture-alone control.
FrozenRandom-GPT2 (standard scaling, GPT-2 init) is the
architecture-alone control reported in §3.2; its NC2 cube-task1 failure
(§3.3) closes the same confound at modality scale.

\textbf{B.3 Full CA R90 control table.}

{\def\LTcaptype{none} 
\begin{longtable}[]{@{}
  >{\raggedright\arraybackslash}p{(\linewidth - 10\tabcolsep) * \real{0.1667}}
  >{\raggedright\arraybackslash}p{(\linewidth - 10\tabcolsep) * \real{0.1667}}
  >{\raggedleft\arraybackslash}p{(\linewidth - 10\tabcolsep) * \real{0.1667}}
  >{\raggedleft\arraybackslash}p{(\linewidth - 10\tabcolsep) * \real{0.1667}}
  >{\raggedleft\arraybackslash}p{(\linewidth - 10\tabcolsep) * \real{0.1667}}
  >{\raggedleft\arraybackslash}p{(\linewidth - 10\tabcolsep) * \real{0.1667}}@{}}
\toprule\noalign{}
\begin{minipage}[b]{\linewidth}\raggedright
Control
\end{minipage} & \begin{minipage}[b]{\linewidth}\raggedright
Backbone
\end{minipage} & \begin{minipage}[b]{\linewidth}\raggedleft
Trainable
\end{minipage} & \begin{minipage}[b]{\linewidth}\raggedleft
L10
\end{minipage} & \begin{minipage}[b]{\linewidth}\raggedleft
L20
\end{minipage} & \begin{minipage}[b]{\linewidth}\raggedleft
L30
\end{minipage} \\
\midrule\noalign{}
\endhead
\bottomrule\noalign{}
\endlastfoot
\textbf{FrozenGemma-L24-29} & Pretrained Gemma L24--L29 & 6.1M &
\textbf{0.000} & \textbf{0.0002} & \textbf{0.257} \\
LSTM-matched & LSTM 864 & 6.1M & 0.0003 & 0.066 & 0.287 \\
LSTM-NTM-small & LSTM 256 (Graves) & 0.56M & 0.241 & 0.335 & 0.403 \\
FrozenGemma-L24-29-random & Random Gemma & 6.1M & 0.278 & 0.362 &
0.481 \\
FrozenRandom-GPT2 s42 & Frozen random transformer, \(1/\sqrt{d_k}\) &
6.1M & 0.422 & 0.459 & 0.474 \\
FrozenRandom-GPT2 s1337 & (replication) & 6.1M & 0.462 & 0.486 &
0.493 \\
Identity & Identity pass-through + LN & 6.1M & 0.484 & 0.496 & 0.496 \\
\end{longtable}
}

\textbf{B.4 Adapter capacity does not affect AR or copy.}

{\def\LTcaptype{none} 
\begin{longtable}[]{@{}lrrr@{}}
\toprule\noalign{}
Configuration & Trainable & Copy L30 & AR L30 \\
\midrule\noalign{}
\endhead
\bottomrule\noalign{}
\endlastfoot
FrozenGemma-L24-29 (FiLM + NTM N=128) & 6.1M & 0.156 & 0.052 \\
NTM N=4 / N=1 & 6.1M & 0.147 / --- & 0.050 / 0.050 \\
FrozenGemma-L24-29-thin (linear, no NTM) & \textasciitilde500K & 0.193 &
0.048 \\
FrozenGemma-L25-27-thin (3 layers + linear) & \textasciitilde113K &
0.225 & 0.057 \\
\end{longtable}
}

NTM memory-slot count does not affect L30. A 113K-parameter linear
interface around three Gemma layers reproduces the 6.1M NTM-pipeline
behavior. Scaling the adapter up (60M transformer adapter) does not
help; scaling it down does not hurt. \textbf{The adapter-capacity result
converts the borrowed-geometry claim from feature-reuse to
computation-reuse.}

\textbf{B.5 Matched-capacity Trained-Transformer per-task detail.}
Trained-Transformer: 6.36M from-scratch pre-LN transformer (2L
\(\times\) \(d=512\), 8 heads, FFN 2048, GELU, \(1/\sqrt{d_k}\), causal
mask) wrapped in NTM(N=128, M=20) + linear decoder. Two seeds, LR sweep
\(\{1\mathrm{e}{-}4, 3\mathrm{e}{-}4\}\).

{\def\LTcaptype{none} 
\begin{longtable}[]{@{}
  >{\raggedright\arraybackslash}p{(\linewidth - 6\tabcolsep) * \real{0.2143}}
  >{\raggedleft\arraybackslash}p{(\linewidth - 6\tabcolsep) * \real{0.2857}}
  >{\raggedleft\arraybackslash}p{(\linewidth - 6\tabcolsep) * \real{0.2857}}
  >{\raggedright\arraybackslash}p{(\linewidth - 6\tabcolsep) * \real{0.2143}}@{}}
\toprule\noalign{}
\begin{minipage}[b]{\linewidth}\raggedright
Task
\end{minipage} & \begin{minipage}[b]{\linewidth}\raggedleft
Best frozen-Gemma
\end{minipage} & \begin{minipage}[b]{\linewidth}\raggedleft
Trained-Transformer best
\end{minipage} & \begin{minipage}[b]{\linewidth}\raggedright
Verdict
\end{minipage} \\
\midrule\noalign{}
\endhead
\bottomrule\noalign{}
\endlastfoot
\textbf{AR L30} & \textbf{0.051} & 0.432 & \textbf{FrozenGemma wins
\(8.7\times\)} (§3.1) \\
\textbf{CA R90 L20} & 0.0001 & 0.0217 & \textbf{FrozenGemma wins
\(217\times\)} \\
CA R90 L30 & 0.257 & 0.164 & Trained-Transformer wins \(1.57\times\) \\
Copy L30 & 0.193 & 0.068 & Trained-Transformer wins \(2.85\times\) \\
Dyck-2 L30 & 0.027 & 0.0011 & Trained-Transformer wins \(25\times\) \\
\end{longtable}
}

AR is the supervised case where the frozen-Gemma transfer-helps result
survives the matched-capacity control.

\textbf{B.6 Activation patching on supervised circuit (negative
result).} We test whether named heads causally transfer the matching
variable on supervised tasks via Wang--IOI-style activation patching.
For each (task, head) we build 192 (clean, corrupt) batches differing in
the matching variable, capture the head's \(W_O\) input on the clean
run, and replace it during corrupt. Recovery rate \(R_{\text{clean}}\)
is the fraction of recall-position bits where the patched run produces
the \emph{clean} answer (chance = 0.5 for binary). Full table in §6.1
main body. Key finding: no head transfers the matching variable in
isolation; named heads ARE causally important (AR L27.2 patching
disrupts \(0.94 \to 0.56\)) but do not individually carry the variable.

\textbf{B.7 Sufficiency: 4-head subsets do not suffice (negative
result).} Mask all-but-\(k\) heads in the L24--L29 slice;
\emph{named\_only} keeps the four §4.3 heads, \emph{random4\_only} keeps
four random non-named, \emph{named\_inverse} zeros only the four named,
\emph{zero\_all} zeros all 192. Full table in §6.2 main body.
\emph{named\_only} sits at chance error on every task, indistinguishable
from random4\_only and zero\_all. \emph{named\_inverse} close to full on
AR (0.07 vs 0.05). Supervised circuit is distributed across the 6L
slice.

\textbf{B.8 Per-length CA R90 + sample efficiency.} Per-length detail
and step-to-cross-0.01 sample-efficiency comparisons (FrozenGemma
\textasciitilde4k vs LSTM-NTM-small \textasciitilde13k on copy L10) in
the project repository.

\subsubsection{Appendix C --- Generalization (full sweep, hypergeometric
sensitivity, permutation
null)}\label{appendix-c-generalization-full-sweep-hypergeometric-sensitivity-permutation-null}

\textbf{C.1 141-pair sweep.}

{\def\LTcaptype{none} 
\begin{longtable}[]{@{}lrr@{}}
\toprule\noalign{}
Task & Heads with \(\Delta > 0.05\) & Fraction of 192 \\
\midrule\noalign{}
\endhead
\bottomrule\noalign{}
\endlastfoot
AR (L30) & 1 & 0.5\% \\
CA R90 (L20) & 28 & 14.6\% \\
Copy (L30) & 42 & 21.9\% \\
Addition (L30) & 70 & 36.5\% \\
\end{longtable}
}

\needspace{23em}

\begin{figure}[H]\centering

\includegraphics[width=1\linewidth,height=\textheight,keepaspectratio,alt={Head-ablation criticality across 192 heads × 4 tasks}]{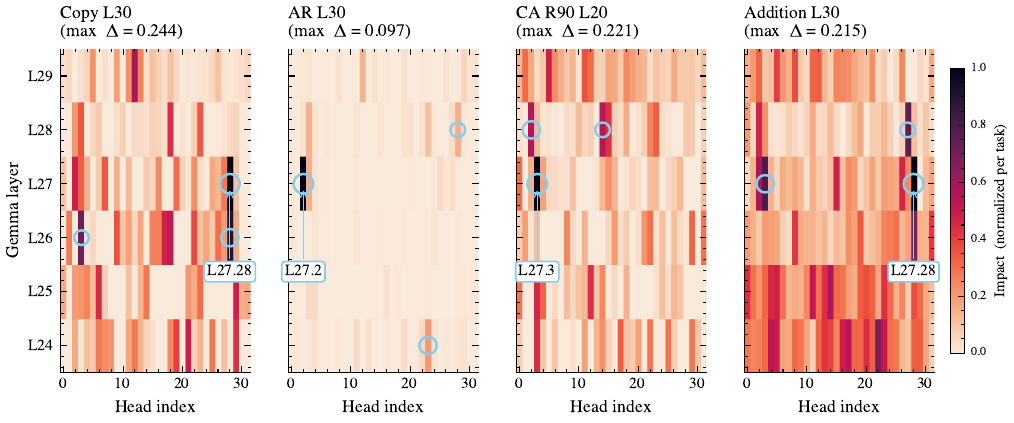}

\caption*{

\textbf{Figure 7. Sparse, task-specific criticality.} \emph{Different
tasks select different dominant heads on the same physical layer L27.}

}

\end{figure}

\textbf{C.2 Single-function classification.} Of 141 critical pairs at
threshold, 67 (47.5\%) classify as single-function above \(1.5\times\)
slice baseline. TxtCopy (24 pairs), Induction (22 pairs), PrevToken (15
pairs) account for 90\% of single-function classifications. Consistent
with the prediction of superposition theory \citep{elhage2022toy}:
features whose pretraining importance permits it crystallize into clean
directions; others remain in superposition.

\textbf{C.3 Hypergeometric sensitivity + permutation null V1--V4.}

{\def\LTcaptype{none} 
\begin{longtable}[]{@{}
  >{\raggedleft\arraybackslash}p{(\linewidth - 4\tabcolsep) * \real{0.3333}}
  >{\raggedleft\arraybackslash}p{(\linewidth - 4\tabcolsep) * \real{0.3333}}
  >{\raggedleft\arraybackslash}p{(\linewidth - 4\tabcolsep) * \real{0.3333}}@{}}
\toprule\noalign{}
\begin{minipage}[b]{\linewidth}\raggedleft
\(K\) (top-\(K\) by TxtCopy)
\end{minipage} & \begin{minipage}[b]{\linewidth}\raggedleft
\(x_{\text{obs}}\) of 4
\end{minipage} & \begin{minipage}[b]{\linewidth}\raggedleft
\(P(X \geq x_{\text{obs}})\)
\end{minipage} \\
\midrule\noalign{}
\endhead
\bottomrule\noalign{}
\endlastfoot
15 & 1 & 0.280 \\
20 & 2 & 0.055 \\
25 & 2 & 0.083 \\
30 & 3 & 0.012 \\
31 & \textbf{4} & \(\mathbf{5.7 \times 10^{-4}}\) \\
35 & 4 & \(9.5 \times 10^{-4}\) \\
\textbf{38} & \textbf{4} & \(\mathbf{0.0013}\) (pre-specified
threshold) \\
40 & 4 & 0.0017 \\
45 & 4 & 0.0027 \\
\end{longtable}
}

\textbf{Permutation null (multiplicity-aware).} To address researcher
degrees of freedom in §4.4, we run permutation tests that hold
ablation-selected heads constant (\(\{\)L26.28, L27.28, L27.2,
L27.3\(\}\)) and permute TxtCopy scores across the 192 slice heads; same
reporting rule, 10K shuffles per variant.

{\def\LTcaptype{none} 
\begin{longtable}[]{@{}llr@{}}
\toprule\noalign{}
Variant & Multiplicity over & \(P_{\text{emp}}\) \\
\midrule\noalign{}
\endhead
\bottomrule\noalign{}
\endlastfoot
V1 & TxtCopy only, \(K=38\) (paper-config) & 0.0013 \\
V2 & best of \(\{\)TxtCopy, Induction, PrevToken\(\}\), \(K=38\) &
0.0013 \\
V3 & TxtCopy, best \(K \in \{30, 38, 50\}\) & 0.0037 \\
\textbf{V4} & \textbf{V2 \(\cup\) V3 (max multiplicity)} &
\textbf{0.0133} \\
\end{longtable}
}

V4 is the conservative bound: even allowing full search over probe
primitive AND threshold \(K\), the joint coincidence remains significant
at \(P=0.013\). We report V4 alongside V1 to address
\citet{wang2022ioi}'s minimality criterion at the level of the
statistical test.

\textbf{C.4 L27 as task-routing layer.} L27 hosts different dominant
heads for different tasks: L27.28 for copy/addition, L27.2 for AR, L27.3
for CA R90.

\textbf{C.5 Full-rank geometry.} Per-head SVD on Q/K/V/O/MLP in
L24--L29: \textasciitilde50\% of singular values needed for 90\%
spectral energy. Truncating output projections to top 256
(\textasciitilde5\%) collapses task performance to random chance across
all four working tasks.

\subsubsection{Appendix D --- Coverage boundaries beyond
Dyck-2}\label{appendix-d-coverage-boundaries-beyond-dyck-2}

\textbf{D.1 GoL ceiling (4-surface).} LSTM-big 0.050, FrozenGemma-L24-29
0.213, HR per-head 0.261, LIMe K/V 0.322, AGCN 0.286 (best L8 Hilbert,
\(n=1\) each). AGCN gives direct access to the substrate's 192
attention-score matrices as graph adjacencies and still fails.

\textbf{D.2 Continuous time series.} NARMA-10 L2000 NRMSE: LSTM 0.34,
ESN 0.85, FrozenGemma 0.87. MG-17 L2000: LSTM 0.003, ESN 0.051,
FrozenGemma 0.122.

\textbf{D.3 D4RL HalfCheetah / Hopper.} HalfCheetah-medium-v2: Gemma-DT
37.7 vs DT 42.6 / IQL 47.4. Hopper-medium-v2: Gemma-DT peak 47.7 at iter
30k followed by collapse; DT 67.6, IQL 66.3. Borrowed geometry does not
provide fine-continuous-control precision or compounding-error recovery.

\textbf{D.4 Addition carries.} Best-checkpoint \(n=3\): LSTM-NTM-small
mean L30 = 0.195 vs FrozenGemma-L24-29 mean L30 = 0.209 --- LSTM beats
by 7\%. Earlier ``geometry wins addition'' claim was a final-step
artifact at smaller \(n\).

\textbf{D.5 Memorization ceilings.} Sokoban-Boxoban BC action accuracy:
FrozenGemma-L24-29-thin 60\% vs LSTM-NTM-small 30\% (\(4\times\)
speedup); held-out solve rate 0\% across all architectures. MiniGrid
MultiRoom-N4-S5: 10-envs, LSTM-NTM-small 27.4\% vs FrozenGemma 26.0\%.
Pong-RAM BC: \textasciitilde74\% action-accuracy ceiling across
LSTM-NTM-small (6.2M), FrozenGemma-L24-29-thin (742K),
Trained-Transformer (30M).

\subsubsection{Appendix E --- Distillation predicts circuit
layout}\label{appendix-e-distillation-predicts-circuit-layout}

\textbf{E.1 Setup.} Student: 4-layer pre-LN decoder transformer,
hidden=256, 5.27M trainable. Teacher: FrozenGemma-L24-29-thin pipeline.
Three aux variants: DS-MSE (learnable \(5376{\times}256\) up-projection
+ L2), DS-CKA (unbiased HSIC CKA), DS-PROC (orthogonal Procrustes
residual via fp64 SVD). Hint pairing: single = student final
\(\leftrightarrow\) Gemma L29; multi = student L1\(\leftrightarrow\)G24,
L2\(\leftrightarrow\)G26, L3\(\leftrightarrow\)G28,
L4\(\leftrightarrow\)G29. Three-phase schedule with
\(\lambda_{\text{aux}} = 0.1\) at fine-tune.

\textbf{E.2 Associative recall.}

{\def\LTcaptype{none} 
\begin{longtable}[]{@{}
  >{\raggedright\arraybackslash}p{(\linewidth - 10\tabcolsep) * \real{0.1304}}
  >{\raggedleft\arraybackslash}p{(\linewidth - 10\tabcolsep) * \real{0.1739}}
  >{\raggedleft\arraybackslash}p{(\linewidth - 10\tabcolsep) * \real{0.1739}}
  >{\raggedleft\arraybackslash}p{(\linewidth - 10\tabcolsep) * \real{0.1739}}
  >{\raggedleft\arraybackslash}p{(\linewidth - 10\tabcolsep) * \real{0.1739}}
  >{\raggedleft\arraybackslash}p{(\linewidth - 10\tabcolsep) * \real{0.1739}}@{}}
\toprule\noalign{}
\begin{minipage}[b]{\linewidth}\raggedright
Condition
\end{minipage} & \begin{minipage}[b]{\linewidth}\raggedleft
L4
\end{minipage} & \begin{minipage}[b]{\linewidth}\raggedleft
L8
\end{minipage} & \begin{minipage}[b]{\linewidth}\raggedleft
L12
\end{minipage} & \begin{minipage}[b]{\linewidth}\raggedleft
L20
\end{minipage} & \begin{minipage}[b]{\linewidth}\raggedleft
L30
\end{minipage} \\
\midrule\noalign{}
\endhead
\bottomrule\noalign{}
\endlastfoot
Student-only & 0.068 & 0.246 & 0.334 & 0.377 & 0.408 \\
DS-CKA single-hint & 0.356 & 0.397 & 0.405 & 0.415 & 0.443 (harms) \\
DS-PROC single-hint & 0.007 & 0.063 & 0.124 & 0.255 & 0.331 \\
DS-MSE single-hint & 0.0005 & 0.013 & 0.023 & 0.036 & 0.094 \\
DS-CKA multi-hint & 0.002 & 0.009 & 0.019 & 0.040 & 0.106 \\
\textbf{DS-PROC multi-hint (s42)} & \textbf{0.002} & \textbf{0.008} &
\textbf{0.015} & \textbf{0.029} & \textbf{0.051} \\
FrozenGemma-L25-27-thin reference & 0.0015 & 0.0073 & 0.0127 & 0.0195 &
0.0415 \\
\end{longtable}
}

DS-PROC multi-hint at s42 closes AR to within 24\% of Gemma online (L30
0.0513 vs 0.0415, \(1.24\times\)) at zero Gemma forward at inference.
\textbf{Circuit-layout-predicts-strategy:} §4.3 shows AR is distributed
across L26/L27/L29; multi-hint required.

\needspace{28em}

\begin{figure}[H]\centering

\includegraphics[width=0.85\linewidth,height=\textheight,keepaspectratio,alt={Associative recall --- distillation results}]{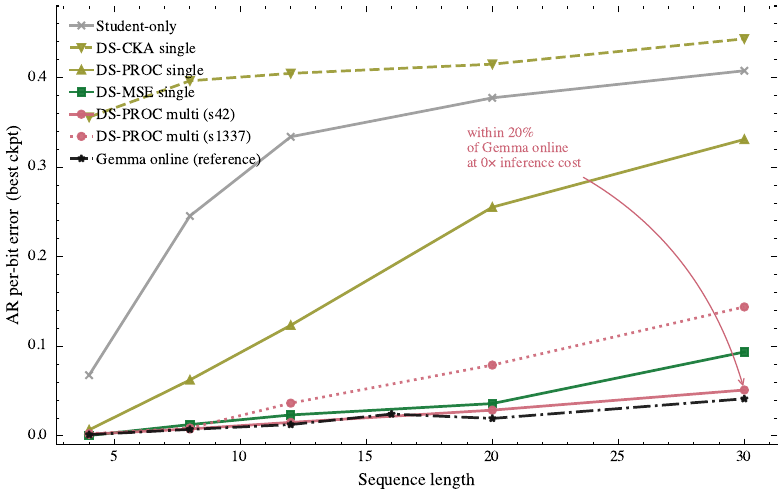}

\caption*{

\textbf{Figure 8. Distillation circuit-prediction.} \emph{AR per-bit
error vs length. DS-PROC multi-hint at s42 closes to within 24\% of
online Gemma at zero Gemma inference cost.}

}

\end{figure}

\textbf{E.3 Other tasks.} Copy (localized at L27.28): DS-PROC
single-hint L30 = 0.030 (\(1.2\times\) over Student-only); multi-hint
does not improve. CA R90: DS-PROC multi-hint L30 = 0.138 vs Student-only
0.171 (\(-19\%\)). Addition: DS-PROC L30 = 0.067 vs Student-only 0.085
(\(-21\%\)).

\textbf{E.4 Distill-embed contrastive.} A 27.5M-parameter bidirectional
student trained on MS-MARCO BM25-50 triples with DS-PROC multi-hint
reaches training-set probe InfoNCE 0.087 vs batch-softmax baseline 0.189
(\(2.2\times\) lower). MTEB transfer pre-registered.

\subsubsection{Appendix F --- Cross-modality results outside the
named-head
triple}\label{appendix-f-cross-modality-results-outside-the-named-head-triple}

\textbf{F.1 D4RL Walker2d-medium-v2 --- DT parity.} Gemma-DT replaces
DT's 3-layer GPT-2 body with frozen Gemma L24--L29. Trainable interface:
per-modality linear embeddings + position embedding + action head +
LayerNorm + input-statistics matcher (\textasciitilde521K trainable).
Best-checkpoint \(n=3\): 76.2 ± 0.8 normalized; all three seeds exceed
Chen 2021 DT (74.0, \textasciitilde1.2M trainable) at \(0.43\times\)
DT's count.

Layer-drop sweep (\(n=3\)):

{\def\LTcaptype{none} 
\begin{longtable}[]{@{}
  >{\raggedright\arraybackslash}p{(\linewidth - 10\tabcolsep) * \real{0.1304}}
  >{\raggedleft\arraybackslash}p{(\linewidth - 10\tabcolsep) * \real{0.1739}}
  >{\raggedleft\arraybackslash}p{(\linewidth - 10\tabcolsep) * \real{0.1739}}
  >{\raggedleft\arraybackslash}p{(\linewidth - 10\tabcolsep) * \real{0.1739}}
  >{\raggedleft\arraybackslash}p{(\linewidth - 10\tabcolsep) * \real{0.1739}}
  >{\raggedleft\arraybackslash}p{(\linewidth - 10\tabcolsep) * \real{0.1739}}@{}}
\toprule\noalign{}
\begin{minipage}[b]{\linewidth}\raggedright
Frozen slice
\end{minipage} & \begin{minipage}[b]{\linewidth}\raggedleft
Frozen params
\end{minipage} & \begin{minipage}[b]{\linewidth}\raggedleft
\(n\)
\end{minipage} & \begin{minipage}[b]{\linewidth}\raggedleft
Mean
\end{minipage} & \begin{minipage}[b]{\linewidth}\raggedleft
Std
\end{minipage} & \begin{minipage}[b]{\linewidth}\raggedleft
\(\Delta\) vs 6L
\end{minipage} \\
\midrule\noalign{}
\endhead
\bottomrule\noalign{}
\endlastfoot
6L L24--L29 (baseline) & 2.93B & 3 & 76.18 & 0.79 & --- \\
\textbf{5L L25--L29 (drop L24)} & 2.45B & 3 & \textbf{77.84} &
\textbf{0.62} & \textbf{+1.66} \\
3L L26--L28 (named-head core) & 1.46B & 3 & 74.52 & 2.31 & \(-1.66\) \\
2L L26--L27 & 977M & 3 & 75.58 & 3.16 & \(-0.60\) \\
1L L27 & 488M & 3 & 65.89 & 18.30 & \(-10.29\) \\
1L L26 & 488M & 3 & 65.88 & 15.86 & \(-10.30\) \\
\textbf{1L L24 (edge-of-slice control)} & 488M & 3 & \textbf{73.81} &
8.04 & \(-2.37\) \\
\end{longtable}
}

L24 (edge-of-slice control, outside the §4.3 named triple) is the most
robust single-layer slice. 1L\_27 (§4.3 universal-head layer on
supervised tasks) collapses on one of three seeds. \textbf{Walker2d
transfer does not localize to §4.3 named heads.}

\textbf{Walker2d-DT head ablation table (\(n=30\), paired env-reset
seeds).} Following the cube-task1 protocol, we zero one head at a time
in the trained 5L slice and re-evaluate. Recruited layer (L24, per
layer-drop sweep above) is OUTSIDE this 5L (L25--L29) checkpoint.

{\def\LTcaptype{none} 
\begin{longtable}[]{@{}lrr@{}}
\toprule\noalign{}
Condition & Norm score & \(\Delta\) from baseline \\
\midrule\noalign{}
\endhead
\bottomrule\noalign{}
\endlastfoot
Baseline (no ablation) & 69.86 & --- \\
L27.28 NAMED & 63.45 & \(-6.41\) \\
L26.28 NAMED & 52.03 & \(-17.83\) \\
L27.3 NAMED & 47.71 & \(-22.15\) \\
L27.2 NAMED & 45.00 & \(-24.86\) \\
L26.5 NEG & 44.60 & \(-25.26\) \\
L27.15 NEG & 50.28 & \(-19.58\) \\
\end{longtable}
}

Named-head drops (\(-6\) to \(-25\)pt) are statistically
indistinguishable from neg-control drops (\(-20\) to \(-25\)pt). Compare
with cube-task1 §5 where L26.28 drops 53pt vs L26.5 drops 17pt
(\(3.2\times\)). The §4 mechanism is layer-localized; Walker2d's
recruited L24 sits outside this slice. P1 prediction: confirmed in both
directions (cube collapse, walker null).

\needspace{27em}

\begin{figure}[H]\centering

\includegraphics[width=0.85\linewidth,height=\textheight,keepaspectratio,alt={Walker2d-medium-v2 --- n=3}]{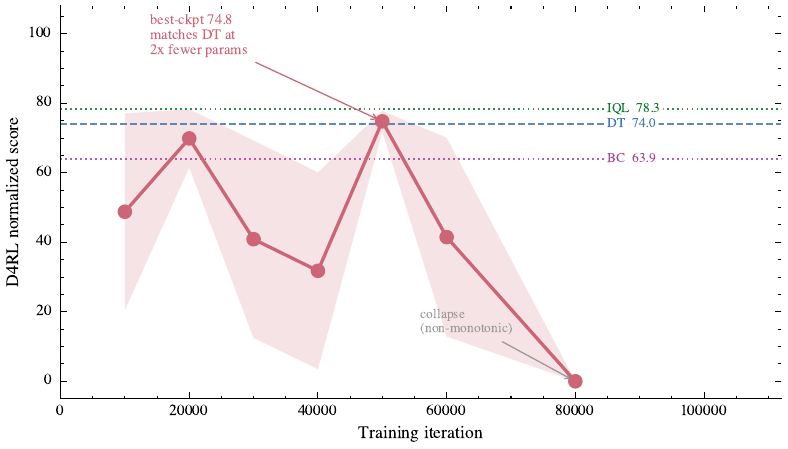}

\caption*{

\textbf{Figure 9. Walker2d \(n=3\) result.} \emph{Gemma-DT D4RL
normalized score: per-seed best-checkpoint trajectories. All three seeds
reach DT-parity. Reference lines: BC (63.9), DT 1.2M (74.0), IQL (78.3).
Trainable: 521K.}

}

\end{figure}

\textbf{F.2 OGBench scene-play-task1 --- SOTA win at \(n=3\).} GemmaIQL
on a single layer L24 (488M frozen). \(n=3\) seeds, last-3-mean =
97.33\% ± 0.74 vs published GCIQL = 93\%, \(\Delta = +4.33\)pt.~All
three seeds above 96\%. NC1 random-Gemma L24 = 86.89\% ± 8.70
(\(\sim 10\times\) wider std); pretrained substrate's primary signal on
saturated tasks is \emph{training stability}, not mean gap. L24 outside
the §4.3 named-head triple.

\textbf{F.3 Cube-task1 L26 seed-fragility detail.} The \(n=3\) L26 mean
of 60.22\% has std 45.87 because s2024 collapses while s42 + s1337 reach
88--96\%. Fig 14b shows the per-seed training curves: s42 climbs above
GCIQL early; s1337 reaches GCIQL at iter 80K; s2024 peaks at 24\% at
iter 70K and collapses to 0\% by iter 99K.

\needspace{27em}

\begin{figure}[H]\centering

\includegraphics[width=0.85\linewidth,height=\textheight,keepaspectratio,alt={L26 cube-task1 individual seed training curves}]{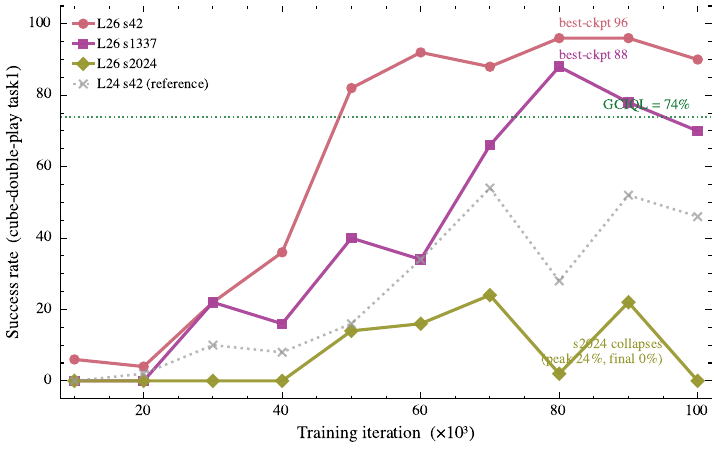}

\caption*{

\textbf{Figure 10. L26 cube-task1 per-seed curves.}

}

\end{figure}

\textbf{F.4 Multi-attribute matched controls and within-layer regression
on L26 (M2).} To test whether L26.28's \(-53.3\)pt drop is explained by
simpler properties, we computed three attribute statistics on cube-task1
observations (\(n=256\) states): (a) output-projection weight norm
\(\lVert W_O[:, h\!\cdot\! d_h:(h+1)d_h] \rVert_F\), (b) per-head
activation norm \(\lVert \mathrm{attn\_out}_h \rVert_2\), (c) per-head
activation CV. We matched each axis to L26.28 by closest value (top-3
nearest, excluding L26.28) and read off ablation drops:

{\def\LTcaptype{none} 
\begin{longtable}[]{@{}
  >{\raggedright\arraybackslash}p{(\linewidth - 8\tabcolsep) * \real{0.1667}}
  >{\raggedright\arraybackslash}p{(\linewidth - 8\tabcolsep) * \real{0.1667}}
  >{\raggedleft\arraybackslash}p{(\linewidth - 8\tabcolsep) * \real{0.2222}}
  >{\raggedleft\arraybackslash}p{(\linewidth - 8\tabcolsep) * \real{0.2222}}
  >{\raggedleft\arraybackslash}p{(\linewidth - 8\tabcolsep) * \real{0.2222}}@{}}
\toprule\noalign{}
\begin{minipage}[b]{\linewidth}\raggedright
Match axis
\end{minipage} & \begin{minipage}[b]{\linewidth}\raggedright
Matched heads
\end{minipage} & \begin{minipage}[b]{\linewidth}\raggedleft
Mean drop (matched)
\end{minipage} & \begin{minipage}[b]{\linewidth}\raggedleft
L26.28 drop
\end{minipage} & \begin{minipage}[b]{\linewidth}\raggedleft
Ratio
\end{minipage} \\
\midrule\noalign{}
\endhead
\bottomrule\noalign{}
\endlastfoot
output-proj norm & L26.\{25, 18, 22\} & \(-45.6\)pp & \(-53.3\)pp &
\(1.2\times\) \\
activation norm & L26.\{19, 8, 5\} & \(-18.9\)pp & \(-53.3\)pp &
\(\mathbf{2.8\times}\) \\
activation CV & L26.\{8, 5, 19\} & \(-18.9\)pp & \(-53.3\)pp &
\(\mathbf{2.8\times}\) \\
original NEG & L26.5 & \(-16.7\)pp & \(-53.3\)pp &
\(\mathbf{3.2\times}\) \\
\end{longtable}
}

L26.28 is significantly more impactful than activation-magnitude-matched
and activation-CV-matched controls; the output-projection-norm-matched
comparison narrows to \(1.2\times\), indicating \(W_O\) norm is
partially predictive of cube-task1 ablation impact within L26.

\textbf{Multivariate regression: TxtCopy does not survive within-layer
controls.} Regressing ablation drop (pp) on standardized (TxtCopy ratio,
\(W_O\) norm, activation norm, activation CV) over all 32 L26 heads
gives \(R^2 = 0.15\), no predictor reaching \(p < 0.05\) in the joint
model (TxtCopy: \(p = 0.08\), \(W_O\) norm: \(p = 0.75\), activation
norm: \(p = 0.18\), activation CV: \(p = 0.32\)). Univariate Spearman:
\(\rho(\text{TxtCopy, drop}) = +0.37\) (\(p = 0.039\), \(n=32\)) --- the
\emph{opposite} direction of a within-layer causal reading. The §4.3
dual-measurement coincidence is therefore a \emph{slice-level enrichment
pattern} (high-TxtCopy heads cluster in the top-K of the 192-head
ablation distribution across four supervised tasks, App C.3) and not a
within-layer prediction of which L26 head will dominate cube-task1
ablation impact.

\textbf{F.5 Standalone-student distillation on Walker2d (\(n=1\) in
flight).} 28M-trainable 4-layer student distilled from the pretrained 5L
Gemma slice (F.1) reaches Walker2d-medium-v2 normalized 79.97 at zero
Gemma at inference (s42, \(n=1\); \(n=3\) replication in flight).
Negative controls: random-teacher distill 0.06; student-only 0.04;
random-Gemma 5L 63.09. Substrate-isolation \(+15.10\)pt at \(n=3\).

\subsubsection{Appendix G --- Extraction
surfaces}\label{appendix-g-extraction-surfaces}

{\def\LTcaptype{none} 
\begin{longtable}[]{@{}
  >{\raggedright\arraybackslash}p{(\linewidth - 4\tabcolsep) * \real{0.3333}}
  >{\raggedright\arraybackslash}p{(\linewidth - 4\tabcolsep) * \real{0.3333}}
  >{\raggedright\arraybackslash}p{(\linewidth - 4\tabcolsep) * \real{0.3333}}@{}}
\toprule\noalign{}
\begin{minipage}[b]{\linewidth}\raggedright
Surface
\end{minipage} & \begin{minipage}[b]{\linewidth}\raggedright
Reads
\end{minipage} & \begin{minipage}[b]{\linewidth}\raggedright
Key finding
\end{minipage} \\
\midrule\noalign{}
\endhead
\bottomrule\noalign{}
\endlastfoot
\textbf{Residual} (canonical) & Post-L29 residual & Baseline; wins on
working tasks \\
\textbf{Per-head} (HR) & 192 per-head outputs + gate & Task-specific
gate patterns localize roughly to §4.3 named heads \\
\textbf{Multi-layer K/V} (LIMe) & K/V from each of 6 layers & Dyck-2
plateau holds; L29 K/V dominates \\
\textbf{Attention adjacency} (AGCN) & 192 attention-score matrices as
graph + GCN & GoL plateau holds; copy L30 = 0.177 (\(n=1\)); only
non-residual to beat canonical thin baseline \\
\textbf{Recurrent depth} & Cycle slice \(K\) times & Dyck plateau holds;
copy destroyed at K=8 \\
\textbf{Body-replacement} (Gemma-DT) & Substrate as sequence-model body
for \((R, s, a)\) tokens & Walker2d transfer; embedding not read-out \\
\end{longtable}
}

KV-cache-memory: \(4\times\) simpler than canonical, matches
FrozenGemma-L24-29 on AR (\(n=1\)).

\subsubsection{Appendix H --- Best-checkpoint protocol
rationale}\label{appendix-h-best-checkpoint-protocol-rationale}

Training loss is decoupled from eval return on offline RL: on Walker2d,
loss \(\approx 0.04\) across all three seeds while eval scores diverged
from 0 to 77 under identical configuration; \texttt{final.pt} varied
0.1--61.9 across same-config seeds. Per-iter checkpointing + online-eval
history mandatory for ceiling recovery. \(n=1\) in offline RL
structurally unreliable. Best-checkpoint protocol applied uniformly.

\subsubsection{Appendix I ---
Reproducibility}\label{appendix-i-reproducibility}

Code, configs, raw \texttt{metrics.json} traces, and per-seed
checkpoints at the project repository. Compute totals: \textasciitilde9
L40S GPU-hours (CA R90 controls) + \textasciitilde6 H100 GPU-hours
(Walker2d distillation) + \textasciitilde24 H100-hours (OGBench) +
\textasciitilde1 H100-hour (cube/walker head-ablation experiments) +
permutation null computed on CPU.

\subsubsection{\texorpdfstring{Appendix J --- Multi-task shared adapter
(\(n=1\)
observation)}{Appendix J --- Multi-task shared adapter (n=1 observation)}}\label{appendix-j-multi-task-shared-adapter-n1-observation}

A single trained interface with 128-param task-ID embedding routes
across four tasks. At matched-exposure budget (12.5k effective
per-task), 3/5 task-length pairs show positive transfer through the
shared interface, 1 ties, 1 favors per-task. \(n=1\) seed; reported as
observation; upgrade requires \(n \geq 3\) replication.

{\def\LTcaptype{none} 
\begin{longtable}[]{@{}
  >{\raggedright\arraybackslash}p{(\linewidth - 6\tabcolsep) * \real{0.2143}}
  >{\raggedleft\arraybackslash}p{(\linewidth - 6\tabcolsep) * \real{0.2857}}
  >{\raggedleft\arraybackslash}p{(\linewidth - 6\tabcolsep) * \real{0.2857}}
  >{\raggedright\arraybackslash}p{(\linewidth - 6\tabcolsep) * \real{0.2143}}@{}}
\toprule\noalign{}
\begin{minipage}[b]{\linewidth}\raggedright
Task / metric
\end{minipage} & \begin{minipage}[b]{\linewidth}\raggedleft
Per-task FrozenGemma-L24-29 (12.5k best)
\end{minipage} & \begin{minipage}[b]{\linewidth}\raggedleft
Multi-task (50k total best)
\end{minipage} & \begin{minipage}[b]{\linewidth}\raggedright
Verdict
\end{minipage} \\
\midrule\noalign{}
\endhead
\bottomrule\noalign{}
\endlastfoot
Copy L30 & 0.188 & 0.225 & per-task \(1.2\times\) \\
AR L30 & 0.051 & 0.051 & tied \\
CA R90 L20 & 0.133 & 0.076 & multi-task \(1.7\times\) \\
CA R90 L30 & 0.325 & 0.234 & multi-task \(1.4\times\) \\
Addition L30 & 0.361 & 0.254 & multi-task \(1.4\times\) \\
\end{longtable}
}

\subsubsection{Appendix K --- Cross-disciplinary
framings}\label{appendix-k-cross-disciplinary-framings}

The deep-and-cheap-learning argument \citep{lin2017why} grounds the
borrowed-geometry program in physics: natural-world functions are
low-order, symmetric, hierarchical; deep nets approximate them
efficiently because the structure of the data permits compositional
decomposition. \emph{Exaptation} \citep{gouldvrba1982} is the structural
analogue: a head shaped by text-copying recruited frozen for analogous
computation in a non-language domain --- feathers exapted from
thermoregulation to flight. Singular Learning Theory
\citep{mehtaschwab2014} provides a possible formalization of the §7
anti-prior account: loss-landscape singularities induced by language
pretraining may be incompatible with the basin a trained transformer
finds for Dyck-2. Distributed representations \citep{rumelhart1986pdp}
and lottery-ticket subnetworks \citep{frankle2019lottery}: the former
visible in §C.5 (high effective rank), the latter structurally adjacent
to borrowed geometry. Cognitive analogy-transfer literature studies the
structural alignment problem of mapping primitives from source to target
domain --- a conceptual cousin at the symbolic level.

\begin{center}\rule{0.5\linewidth}{0.5pt}\end{center}

\subsection{Acknowledgments}\label{acknowledgments}

Independent research project, self-funded compute. AI coding and
research agents (Claude, OpenAI models) were used for literature review,
experiment iteration, code generation, and draft review; all empirical
results and claims were run, verified, and written by the author. Code
and logs are public at the project repository.

\bibliography{references.bib}

\begin{thebibliography}{26}
\providecommand{\natexlab}[1]{#1}
\providecommand{\url}[1]{\texttt{#1}}
\expandafter\ifx\csname urlstyle\endcsname\relax
  \providecommand{\doi}[1]{doi: #1}\else
  \providecommand{\doi}{doi: \begingroup \urlstyle{rm}\Url}\fi

\bibitem[Black et~al.(2024)Black, Brown, Driess, Esmail, Equi, Finn, Fusai,
  Groom, Hausman, Ichter, et~al.]{black2024pi0}
Kevin Black, Noah Brown, Danny Driess, Adnan Esmail, Michael Equi, Chelsea
  Finn, Niccolo Fusai, Lachy Groom, Karol Hausman, Brian Ichter, et~al.
\newblock {$\pi_0$: A Vision-Language-Action Flow Model for General Robot
  Control}.
\newblock arXiv:2410.24164, 2024.

\bibitem[Chen et~al.(2021)Chen, Lu, Rajeswaran, Lee, Grover, Laskin, Abbeel,
  Srinivas, and Mordatch]{chen2021decision}
Lili Chen, Kevin Lu, Aravind Rajeswaran, Kimin Lee, Aditya Grover, Misha
  Laskin, Pieter Abbeel, Aravind Srinivas, and Igor Mordatch.
\newblock Decision transformer: Reinforcement learning via sequence modeling.
\newblock In \emph{Advances in Neural Information Processing Systems}, 2021.

\bibitem[Driess et~al.(2025)Driess, Springenberg, Ichter, Yu, Li-Bell, Pertsch,
  Ren, Walke, Vuong, Shi, and Levine]{driess2025ki}
Danny Driess, Jost~Tobias Springenberg, Brian Ichter, Lili Yu, Adrian Li-Bell,
  Karl Pertsch, Allen~Z. Ren, Homer Walke, Quan Vuong, Lucy~Xiaoyang Shi, and
  Sergey Levine.
\newblock {Knowledge Insulating Vision-Language-Action Models: Train Fast, Run
  Fast, Generalize Better}.
\newblock arXiv:2505.23705, 2025.

\bibitem[Elhage et~al.(2021)Elhage, Nanda, Olsson, Henighan, Joseph, Mann,
  Askell, Bai, Chen, Conerly, et~al.]{elhage2021mathematical}
Nelson Elhage, Neel Nanda, Catherine Olsson, Tom Henighan, Nicholas Joseph, Ben
  Mann, Amanda Askell, Yuntao Bai, Anna Chen, Tom Conerly, et~al.
\newblock A mathematical framework for transformer circuits.
\newblock \emph{Transformer Circuits Thread}, 2021.

\bibitem[Elhage et~al.(2022)Elhage, Hume, Olsson, Schiefer, Henighan,
  et~al.]{elhage2022toy}
Nelson Elhage, Tristan Hume, Catherine Olsson, Nicholas Schiefer, Tom Henighan,
  et~al.
\newblock Toy models of superposition.
\newblock \emph{Transformer Circuits Thread}, 2022.

\bibitem[Frankle and Carbin(2019)]{frankle2019lottery}
Jonathan Frankle and Michael Carbin.
\newblock The lottery ticket hypothesis: Finding sparse, trainable neural
  networks.
\newblock In \emph{International Conference on Learning Representations
  (ICLR)}, 2019.

\bibitem[Fu et~al.(2020)Fu, Kumar, Nachum, Tucker, and Levine]{fu2024d4rl}
Justin Fu, Aviral Kumar, Ofir Nachum, George Tucker, and Sergey Levine.
\newblock D4rl: Datasets for deep data-driven reinforcement learning.
\newblock In \emph{arXiv preprint arXiv:2004.07219}, 2020.

\bibitem[{Gemma Team}(2026)]{gemmateam2026gemma4}
{Gemma Team}.
\newblock Gemma 4: Frontier multimodal intelligence on device.
\newblock Google DeepMind. \url{https://deepmind.google/models/gemma/gemma-4/},
  2026.
\newblock Released April 2026. Open weights under Apache 2.0.

\bibitem[Gould and Vrba(1982)]{gouldvrba1982}
Stephen~Jay Gould and Elisabeth~S. Vrba.
\newblock Exaptation---a missing term in the science of form.
\newblock \emph{Paleobiology}, 8\penalty0 (1):\penalty0 4--15, 1982.

\bibitem[Graves et~al.(2014)Graves, Wayne, and Danihelka]{graves2014ntm}
Alex Graves, Greg Wayne, and Ivo Danihelka.
\newblock Neural turing machines.
\newblock \emph{arXiv preprint arXiv:1410.5401}, 2014.

\bibitem[Huh et~al.(2024)Huh, Cheung, Wang, and Isola]{huh2024platonic}
Minyoung Huh, Brian Cheung, Tongzhou Wang, and Phillip Isola.
\newblock The platonic representation hypothesis.
\newblock In \emph{International Conference on Machine Learning (ICML)}, 2024.
\newblock arXiv:2405.07987.

\bibitem[Jaeger(2001)]{jaeger2001echo}
Herbert Jaeger.
\newblock The echo state approach to analysing and training recurrent neural
  networks.
\newblock \emph{GMD Report 148, German National Research Center for Information
  Technology}, 2001.

\bibitem[Kim et~al.(2024)Kim, Pertsch, Karamcheti, Xiao, Balakrishna, Nair,
  Rafailov, Foster, Lam, Sanketi, Vuong, Kollar, Burchfiel, Tedrake, Sadigh,
  Levine, Liang, and Finn]{kim2024openvla}
Moo~Jin Kim, Karl Pertsch, Siddharth Karamcheti, Ted Xiao, Ashwin Balakrishna,
  Suraj Nair, Rafael Rafailov, Ethan Foster, Grace Lam, Pannag Sanketi, Quan
  Vuong, Thomas Kollar, Benjamin Burchfiel, Russ Tedrake, Dorsa Sadigh, Sergey
  Levine, Percy Liang, and Chelsea Finn.
\newblock {OpenVLA: An Open-Source Vision-Language-Action Model}.
\newblock arXiv:2406.09246, 2024.

\bibitem[Lin et~al.(2017)Lin, Tegmark, and Rolnick]{lin2017why}
Henry~W. Lin, Max Tegmark, and David Rolnick.
\newblock Why does deep and cheap learning work so well?
\newblock \emph{Journal of Statistical Physics}, 168:\penalty0 1223--1247,
  2017.
\newblock arXiv:1608.08225 (2016).

\bibitem[Lu et~al.(2022)Lu, Grover, Abbeel, and Mordatch]{lu2022frozen}
Kevin Lu, Aditya Grover, Pieter Abbeel, and Igor Mordatch.
\newblock Pretrained transformers as universal computation engines.
\newblock In \emph{AAAI Conference on Artificial Intelligence}, 2022.

\bibitem[Maass et~al.(2002)Maass, Natschl{\"a}ger, and
  Markram]{maass2002liquid}
Wolfgang Maass, Thomas Natschl{\"a}ger, and Henry Markram.
\newblock Real-time computing without stable states: A new framework for neural
  computation based on perturbations.
\newblock \emph{Neural Computation}, 14\penalty0 (11):\penalty0 2531--2560,
  2002.

\bibitem[Marks et~al.(2024)Marks, Rager, Michaud, Belinkov, Bau, and
  Mueller]{marks2024sparse}
Samuel Marks, Can Rager, Eric~J. Michaud, Yonatan Belinkov, David Bau, and
  Aaron Mueller.
\newblock {Sparse Feature Circuits: Discovering and Editing Interpretable
  Causal Graphs in Language Models}.
\newblock In \emph{International Conference on Learning Representations
  (ICLR)}, 2024.
\newblock arXiv:2403.19647.

\bibitem[Mehta and Schwab(2014)]{mehtaschwab2014}
Pankaj Mehta and David~J. Schwab.
\newblock An exact mapping between the variational renormalization group and
  deep learning.
\newblock \emph{arXiv preprint arXiv:1410.3831}, 2014.

\bibitem[Naik and Gupta(2021)]{naikgupta2021}
Aakanksha Naik and Vishwa Gupta.
\newblock Adapting pretrained transformers for tasks outside their training
  distribution.
\newblock \emph{arXiv preprint arXiv:2108.05247}, 2021.

\bibitem[Olsson et~al.(2022)Olsson, Elhage, Nanda, Joseph, DasSarma, Henighan,
  Mann, Askell, Bai, Chen, et~al.]{olsson2022induction}
Catherine Olsson, Nelson Elhage, Neel Nanda, Nicholas Joseph, Nova DasSarma,
  Tom Henighan, Ben Mann, Amanda Askell, Yuntao Bai, Anna Chen, et~al.
\newblock In-context learning and induction heads.
\newblock \emph{Transformer Circuits Thread}, 2022.

\bibitem[Park et~al.(2025)Park, Frans, Eysenbach, and Levine]{park2025ogbench}
Seohong Park, Kevin Frans, Benjamin Eysenbach, and Sergey Levine.
\newblock Ogbench: Benchmarking offline goal-conditioned rl.
\newblock In \emph{International Conference on Learning Representations
  (ICLR)}, 2025.
\newblock arXiv:2410.20092.

\bibitem[{Physical Intelligence}(2025)]{pi06}
{Physical Intelligence}.
\newblock {$\pi^{*}_{0.6}$: a VLA That Learns From Experience}.
\newblock arXiv:2511.14759, 2025.

\bibitem[Rumelhart et~al.(1986)Rumelhart, McClelland, and {PDP Research
  Group}]{rumelhart1986pdp}
David~E. Rumelhart, James~L. McClelland, and {PDP Research Group}.
\newblock \emph{Parallel Distributed Processing: Explorations in the
  Microstructure of Cognition}.
\newblock MIT Press, 1986.

\bibitem[Templeton et~al.(2024)Templeton, Conerly, Marcus, Lindsey, Bricken,
  Chen, Pearce, Citro, Ameisen, Jermyn, Olsson, and Olah]{templeton2024scaling}
Adly Templeton, Tom Conerly, Jonathan Marcus, Jack Lindsey, Trenton Bricken,
  Brian Chen, Adam Pearce, Craig Citro, Emmanuel Ameisen, Andy Jermyn,
  Catherine Olsson, and Christopher Olah.
\newblock {Scaling Monosemanticity: Extracting Interpretable Features from
  Claude 3 Sonnet}.
\newblock Transformer Circuits Thread, 2024.
\newblock URL
  \url{https://transformer-circuits.pub/2024/scaling-monosemanticity/}.

\bibitem[Wang et~al.(2022)Wang, Variengien, Conmy, Shlegeris, and
  Steinhardt]{wang2022ioi}
Kevin Wang, Alexandre Variengien, Arthur Conmy, Buck Shlegeris, and Jacob
  Steinhardt.
\newblock {Interpretability in the Wild: a Circuit for Indirect Object
  Identification in GPT-2 small}.
\newblock arXiv:2211.00593, 2022.

\bibitem[Zitkovich et~al.(2023)Zitkovich, Yu, Xu, Xu, Xiao, Xia, Wu, Wohlhart,
  Welker, Wahid, et~al.]{zitkovich2023rt2}
Brianna Zitkovich, Tianhe Yu, Sichun Xu, Peng Xu, Ted Xiao, Fei Xia, Jialin Wu,
  Paul Wohlhart, Stefan Welker, Ayzaan Wahid, et~al.
\newblock {RT-2: Vision-Language-Action Models Transfer Web Knowledge to
  Robotic Control}.
\newblock In \emph{Conference on Robot Learning (CoRL)}, 2023.
\newblock arXiv:2307.15818.

\end{thebibliography}

\end{document}